\documentclass[10pt]{article} 
\usepackage[preprint]{tmlr}

\usepackage{amsmath,amsfonts,bm}

\def\eqref#1{equation~\ref{#1}}

\def\1{\bm{1}}

\DeclareMathAlphabet{\mathsfit}{\encodingdefault}{\sfdefault}{m}{sl}
\SetMathAlphabet{\mathsfit}{bold}{\encodingdefault}{\sfdefault}{bx}{n}

\DeclareMathOperator*{\argmin}{arg\,min}

\usepackage{microtype}
\usepackage{hyperref}
\usepackage{url,graphicx,subfigure}
\usepackage{booktabs}
\usepackage{amsmath}
\usepackage{amssymb}
\usepackage{mathtools}
\usepackage{amsthm}
\usepackage{multirow}
\usepackage{wrapfig}
\usepackage{subcaption}
\usepackage[capitalize,noabbrev]{cleveref}

\newcommand{\std}[1]{\tiny{$\pm$ #1}}

\title{ProFeAT: Projected Feature Adversarial Training for \\Self-Supervised Learning of Robust Representations}

\author{%
  Sravanti Addepalli \thanks{Equal Contribution.  Correspondence to Sravanti Addepalli $<$sravantia@iisc.ac.in$>$, Priyam Dey $<$priyamdey@iisc.ac.in$>$} ~ \quad Priyam Dey \footnotemark[1] ~ \quad R. Venkatesh Babu \\ Vision and AI Lab, Indian Institute of Science, Bangalore}

\def\month{MM}  
\def\year{YYYY} 
\def\openreview{\url{https://openreview.net/forum?id=XXXX}} 

\begin{document}

\maketitle

\begin{abstract}
The need for abundant labelled data in supervised Adversarial Training (AT) has prompted the use of Self-Supervised Learning (SSL) techniques with AT. However, the direct application of existing SSL methods to adversarial training has been sub-optimal due to the increased training complexity of combining SSL with AT. A recent approach DeACL \citep{deacl} mitigates this by utilizing supervision from a standard SSL teacher in a distillation setting, to mimic supervised AT. However, we find that there is still a large performance gap when compared to supervised adversarial training, specifically on larger models. In this work, investigate the key reason for this gap and propose Projected Feature Adversarial Training (ProFeAT) to bridge the same. We show that the sub-optimal distillation performance is a result of mismatch in training objectives of the teacher and student, and propose to use a projection head at the student, that allows it to leverage weak supervision from the teacher while also being able to learn adversarially robust representations that are distinct from the teacher. We further propose appropriate attack and defense losses at the feature and projector, alongside a combination of weak and strong augmentations for the teacher and student respectively, to improve the training data diversity without increasing the training complexity. Through extensive experiments on several benchmark datasets and models, we demonstrate significant improvements in both clean and robust accuracy when compared to existing SSL-AT methods, setting a new state-of-the-art. We further report on-par/ improved performance when compared to TRADES, a popular supervised-AT method.
\end{abstract}

\section{Introduction}
\vspace{-0.1cm}

Deep Neural Networks are known to be vulnerable to crafted imperceptible input-space perturbations known as \emph{Adversarial attacks} \citep{intriguing-iclr-2014}, which can be used to fool classification networks into predicting any desired output, leading to disastrous consequences. Amongst the diverse attempts at improving the adversarial robustness of Deep Networks, Adversarial Training (AT) \citep{madry-iclr-2018,zhang2019theoretically} has been the most successful. This involves the generation of adversarial attacks by maximizing the training loss, and further minimizing the loss on the generated attacks for training. While adversarial training based methods have proved to be robust against various attacks developed over time \citep{carlini2019evaluating,croce2020reliable,sriramanan2020gama}, they require significantly more training data when compared to standard training \citep{schmidt2018adversarially}, incurring a large annotation cost. This motivates the need for self-supervised learning (SSL) of robust representations, followed by lightweight standard training of the classification head. 
Motivated by the success of contrastive learning for standard self-supervised learning \citep{cpc,simclr,moco}, several works have attempted to use contrastive learning for self-supervised adversarial training as well \citep{acl,rocl,advcl}. While this strategy works well in a full network fine-tuning setting, the performance is sub-optimal when the robustly pretrained feature encoder is frozen while training the classification head (linear probing), demonstrating that the representations learned are indeed sub-optimal. A recent work, Decoupled Adversarial Contrastive Learning (DeACL) \citep{deacl}, demonstrated significant improvements in performance and training efficiency by splitting this combined self-supervised adversarial training into two stages; first, where a standard self-supervised model is trained, and second, where this pretrained model is used as a teacher to provide supervision to the adversarially trained student network. Although the performance of this method is on par with supervised adversarial training on small model architectures (ResNet-18), we find that it does not scale to larger models such as WideResNet-34-10, which is widely reported in the adversarial ML literature.

In this work, we aim to bridge the performance gap between self-supervised and supervised adversarial training methods, and improve the scalability of the former to larger model capacities. We utilize the distillation setting discussed above, where a standard self-supervised trained teacher provides supervision to the student. 
In contrast to a typical distillation scenario, the student's objective or its \emph{ideal goal} is not to replicate the teacher, but to leverage weak supervision from the teacher while also learning adversarially robust representations. This involves a trade-off between the sensitivity towards changes that flip the class of an image (for better clean accuracy) and invariance towards imperceptible perturbations that preserve the true class (for adversarial robustness) \citep{tramer2020fundamental}. Towards this, we propose to impose similarity with respect to the teacher in the appropriate dimensions by applying the distillation loss in a \emph{projected space} (output of a projection MLP layer), while enforcing the smoothness-based robustness loss in the \emph{feature space} (output of a backbone/ feature extractor).  
However, we find that enforcing these losses at different layers results in training instability, and thus introduce the complementary loss (clean distillation loss or robustness loss) as a regularizer to improve training stability. We further propose to reuse the pretrained projection layer from the teacher model for better convergence.

In line with the training objective, the adversarial attack used during training aims to find images that maximize the smoothness loss in the feature space, and cause misalignment between the teacher and student in the projected space. Further, since data augmentations are known to increase the training complexity of adversarial training resulting in a drop in performance \citep{deacl,addepalli2022efficient}, we propose to use augmentations such as AutoAugment (or \emph{strong augmentations}) only at the student for better attack diversity, while using spatial transforms such as pad and crop (PC) (or \emph{weak augmentations}) at the teacher. We summarize our contributions below:
\begin{itemize}
    \item We propose Projected Feature Adversarial Training (ProFeAT) - a teacher-student distillation setting for self-supervised adversarial training, where the projection layer of the standard self-supervised pretrained teacher is utilized for student distillation. We further propose appropriate attack and defense losses for training, coupled with a combination of weak and strong augmentations for the teacher and student respectively.
    \item Towards understanding \emph{why} the projector helps, we first show that the compatibility between the training methodology of the teacher and the ideal goals of the student plays a crucial role in the student model performance in distillation. We further show that the use of a projector can alleviate the negative impact of the inherent misalignment of the above.
    \item We demonstrate the effectiveness of the proposed approach on the standard benchmark datasets CIFAR-10 and CIFAR-100. We obtain significant gains of $3.5-8\%$ in clean accuracy and $\sim3\%$ in robust accuracy on larger model capacities (WideResNet-34-10), and improved performance on small model architectures (ResNet-18), while also outperforming TRADES supervised training \citep{zhang2019theoretically} on larger models.
\end{itemize}

\section{Preliminaries}
\label{sec:prelims}
\vspace{-0.1cm}

We consider the problem of self-supervised learning of robust representations, where a self-supervised standard trained teacher model $\mathcal{T}$ is used to provide supervision to a student model $\mathcal{S}$. The feature, projector and linear probe layers of the teacher are denoted as $\mathcal{T}_f$, $\mathcal{T}_p$ and $\mathcal{T}_l$ respectively. A composition of the feature and projector layers of the teacher is denoted as $\mathcal{T}_{fp} = \mathcal{T}_f \circ \mathcal{T}_p$, and a composition of the feature extractor and linear classification layer is denoted as $\mathcal{T}_{fl} = \mathcal{T}_f \circ \mathcal{T}_l$. An analogous notation is followed for the student as well. 

The dataset used for self-supervised pretraining $\mathcal{D}$ consists of images $x_i$ where $i \leq N$. An adversarial image corresponding to the image $x_i$ is denoted as $\tilde{x}_i$. We consider the $\ell_\infty$ based threat model where $||\tilde{x}_i - x_i||_\infty \leq \varepsilon$. The value of $\varepsilon$ is set to $8/255$ for CIFAR-10 and CIFAR-100 \citep{krizhevsky2009learning}, as is standard in literature \citep{madry-iclr-2018,zhang2019theoretically}. 

To evaluate the representations learned after self-supervised adversarial pretraining, we freeze the pretrained backbone, and perform linear layer training on a downstream labeled dataset consisting of image-label pairs, popularly referred to as linear probing \citep{kumar2022fine}. The training is done using cross-entropy loss on clean samples unless specified otherwise. We compare the robustness of the representations on both in-distribution data, where the linear probing is done using the same distribution of images as that used for pretraining, and in a transfer learning setting, where the distribution of images in the downstream dataset is different from that used for pretraining. We do not consider the case of fine-tuning the full network using adversarial training for our primary evaluations despite its practical relevance, since this changes the pretrained network to large extent, and may yield misleading results and conclusions depending on the dynamics of training (number of epochs, learning rate, and the value of the robustness-accuracy trade-off parameter). Contrary to this, linear probing based evaluation gives an accurate comparison of representations learned across different pretraining algorithms. For the sake of completeness, we present kNN evaluations and Adversarial Full-finetuning based results additionally. 

\vspace{-0.2cm}
\section{Related Works}
\vspace{-0.1cm}

\textbf{Self Supervised Learning (SSL):} 
With the abundance of unlabelled data, learning representations through self-supervision has seen major advances in recent years. 
Early works on self-supervised learning (SSL) had focused on designing well-posed tasks called ``pretext tasks'', such as rotation prediction \citep{rotnet} and solving jigsaw puzzles \citep{jigsaw}, to provide a supervisory signal. However, the design of these hand-crafted tasks involves manual effort, and is generally specific to the dataset and training task. To alleviate this problem, contrastive learning based SSL approaches have emerged as a promising direction \citep{cpc,simclr,moco}, where different augmentations of a given anchor image form positives, and augmentations of other images in the batch form the negatives. The training objective involves pulling the representations of the positives together, and repelling the representations of negatives. Strong augmentations like random cropping and color jitter are applied to make the learning task sufficiently hard, while also enabling the learning of invariant representations.

\textbf{Self Supervised Adversarial Training:} To alleviate the large sample complexity and training cost of adversarial training, there have been several works that have attempted self-supervised learning of adversarially robust representations. 
\citet{chen2020adversarial} propose AP-DPE, an ensemble adversarial pretraining framework where several pretext tasks like Jigsaw puzzles \citep{jigsaw}, rotation prediction \citep{rotnet} and Selfie \citep{selfie} are combined to learn robust representations without task labels. 
\citet{acl} propose ACL, that combines the popular contrastive SSL method - SimCLR \citep{simclr} with adversarial training, 
using Dual Batch normalization layers for the student model - one for the standard branch and another for the adversarial branch. RoCL \citep{rocl} follows a similar approach to ACL by combining the contrastive objective with adversarial training to learn robust representations. \citet{advcl} propose AdvCL, that uses high-frequency components in data as augmentations in contrastive learning, performs attacks on unaugmented images, and uses a pseudo label based loss for training to minimize the cross-task robustness transferability. \citet{dynacl} study the role of augmentation strength in self-supervised contrastive adversarial training, and propose DynACL, that uses a ``strong-to-weak" annealing schedule on augmentations. Additionally, motivated by \citet{kumar2022fine}, they propose DynACL++ that obtains pseudo-labels via k-means clustering on the clean branch of the DynACL pretrained network, and performs linear-probing (LP) using these pseudo-labels followed by adversarial full-finetuning (AFT) of the backbone. This is a generic strategy that can be integrated with several algorithms including ours. 
While most self-supervised adversarial training methods aimed at integrating contrastive learning methods with adversarial training, \citet{deacl} showed that combining the two is a complex optimization problem due to their conflicting requirements. The authors propose Decoupled Adversarial Contrastive Learning (DeACL), where a teacher model is first trained using existing self-supervised training methods such as SimCLR, and further, a student model is trained to be adversarially robust using supervision from the teacher. While existing methods used $\sim1000$ epochs for contrastive adversarial training, the compute requirement for DeACL is much lesser since the first stage does not involve adversarial training, and the second stage is similar in complexity to supervised adversarial training (Details in \cref{app:computational complexity}). We utilize this distillation framework and obtain significant gains over DeACL, specifically for larger models.

\begin{table}[t]
\centering
\caption{\textbf{Diverse representations of Standard Trained (ST) and Adversarially Trained (AT) models (CIFAR-100, WRN-34-10):} ST models achieve 0\% robust accuracy even with adversarial training of the linear layer, and AT models lose their robustness with standard full-finetuning. \textcolor{black}{SA: Standard Accuracy, RA-G: Robust accuracy against GAMA \cite{sriramanan2020gama}, RA-PGD20: Robust Accuracy against PGD-20 \cite{madry-iclr-2018} attack.}}
\resizebox{1.0\linewidth}{!}{
\label{tab:motiv-LP}
\begin{tabular}{lccc|lccc}
\toprule
\multicolumn{1}{c}{\textbf{Training/ LP Method}} & \textbf{SA} & \textbf{\textcolor{black}{RA-PGD20}}      & \textbf{\textcolor{black}{RA-G}}        & \multicolumn{1}{c}{\textbf{Training/ LP Method}} & \textbf{SA} & \textbf{\textcolor{black}{RA-PGD20}}      & \textbf{\textcolor{black}{RA-G}}        \\
\midrule
Standard trained model                           & 80.86                   & 0.00                 & 0.00                 & TRADES AT model                                  & 60.22                   &      28.67                & 26.36                \\
+ Adversarial Linear Probing            &         80.10& 0.00& 0.00 & + Standard Full Finetuning                       &76.11&0.37& 0.11\\
\bottomrule
\end{tabular}}
\end{table}

\section{Proposed Method}
\label{sec:proposed_method}
\vspace{-0.2cm}
We first motivate the need for a projection layer, and further present the proposed approach ProFeAT. 
\vspace{-0.2cm}
\subsection{Projection Layer in Self-supervised Distillation} 
\vspace{-0.2cm}

\label{subsec:motiv-proj}

\begin{table*}
\centering
\caption{\textbf{Role of projector in self-supervised distillation (CIFAR-100, WRN-34-10):} The drop in accuracy of student ($\mathcal{S}$) w.r.t. the teacher  ($\mathcal{T}$) indicates distillation performance, which improves by matching the training objective of the teacher with ideal goals of the student (S3/ S4 vs. S1), and by using similar losses for pretraining and linear probing (LP) (S2 vs. S1). Using a projector improves performance in case of mismatch in the above (S5 vs. S1). The similarity between teacher and student is significantly higher at the projector space when compared to the feature space in S5.}
\resizebox{1.0\linewidth}{!}{
\label{tab:motiv-SD}
\begin{tabular}{clccccccc}
\toprule
\multirow{2}{*}{\textbf{Exp \#}} & \multirow{2}{*}{\textbf{\begin{tabular}[c]{@{}c@{}}Teacher \\ training\end{tabular}}} & \multirow{2}{*}{\textbf{\begin{tabular}[c]{@{}c@{}}Teacher \\ acc (\%)\end{tabular}}} & \multirow{2}{*}{\textbf{Projector}} & \multirow{2}{*}{\textbf{LP Loss}} & \multicolumn{2}{c}{\textbf{Student accuracy after linear probe}} & \multicolumn{2}{c}{\textbf{$\boldsymbol{\cos(\mathcal{T},\mathcal{S})}$}} \\
                                 &                                            &                                                                                       &                                     &                                   & \textbf{\small{Feature space (\%)}}     & \textbf{\small{Projector space (\%)}}    & \textbf{\small{Feature space}}       & \textbf{\small{Projector space}}      \\
                                 \midrule
S1                               & Self-supervised                            & 70.85                                                                                 & Absent                              & CE                                & 64.90                      & -                           & 0.94                         & -                             \\
S2                               & Self-supervised                            & 70.85                                                                                 & Absent                              & $\cos(\mathcal{T},\mathcal{S})$                          & 68.49                      & -                           & 0.94                         & -                             \\
S3                               & Supervised                                 & 80.86                                                                                 & Absent                              & CE                                & 80.40                      & -                           & 0.94                         & -                             \\
S4                               & Supervised                                 & 69.96                                                                                 & Absent                              & CE                                & 71.73                      & -                           & 0.98                         &                               \\
S5                               & Self-supervised                            & 70.85                                                                                 & Present                             & CE                                & 73.14                      & 64.67                       & 0.19                         & 0.92         \\
\bottomrule
\end{tabular}}
\vspace{-0.3cm}
\end{table*}

In this work, we follow the setting proposed by \citet{deacl}, where a standard self-supervised pretrained teacher provides supervision for self-supervised adversarial training of the student model. This is different from a standard distillation setting \citep{hinton2015distilling} because the representations of standard and adversarially trained models are known to be inherently different \citep{engstrom2019adversarial}. \citet{ilyas2019adversarial} attribute the adversarial vulnerability of models to the presence of non-robust features which can be disentangled from robust features that are learned by adversarially trained models. The differences in representations of standard and adversarially trained models can also be justified by the fact that linear probing of standard trained models using adversarial training cannot produce robust models as shown in \cref{tab:motiv-LP}. Similarly, standard full finetuning of adversarially trained models destroys the robust features learned \citep{chen2020adversarial,rocl,advcl}, yielding $0\%$ robustness, as shown in the table. 
Due to the inherently diverse representations of these models, the ideal goal of the student in the considered distillation setting is not to merely follow the teacher, but to be able to take weak supervision from it while being able to differ considerably. In order to achieve this, we take inspiration from standard self-supervised learning literature \citep{cpc,simclr,moco,navaneet2022simreg,disco} and propose to utilize a projection layer following the student backbone, so as to isolate the impact of the enforced loss on the learned representations. \citet{bordes2022guillotine} show that in standard supervised and self-supervised training, a projector is useful when there is a misalignment between the pretraining and downstream tasks, and aligning them can eliminate the need for the same. 
Motivated by this, we hypothesize the following for the setting of self-supervised distillation:

\begin{table*}
\centering
\caption{\textbf{Role of projector in self-supervised adversarial distillation (CIFAR-100, WRN-34-10):} Student performance after linear probe at feature space is reported. The drop in standard accuracy (SA) of the student ($\mathcal{S}$) w.r.t. the teacher  ($\mathcal{T}$), and the robust accuracy (\textcolor{black}{RA-G}) of the student improve by matching the training objective of the teacher with ideal goals of the student (A3 vs. A1), and by using similar losses for pretraining and linear probing (LP) (A2 vs. A1). Using a projector improves
performance in case of mismatch in the above (A4 vs. A1).}
\resizebox{1.0\linewidth}{!}{
\label{tab:motiv-AD}

\begin{tabular}{clccccccc}
\toprule
\multirow{2}{*}{\textbf{Exp \#}} & \multirow{2}{*}{\textbf{Teacher training}}                                     & \multicolumn{2}{c}{\textbf{Teacher accuracy}} & \multirow{2}{*}{\textbf{Projector}} & \multirow{2}{*}{\textbf{LP Loss}}                &  \multicolumn{2}{c}{\textbf{Student accuracy}} & \multirow{2}{*}{\textbf{$\boldsymbol{\cos(\mathcal{T},\mathcal{S})}$}} \\

 && \textbf{SA (\%)} &\textbf{{RA-G} (\%)}&&& \textbf{SA (\%)} & \textbf{{RA-G} (\%)} & \\
\midrule
A1                                      & Self-supervised (standard training)                          & 70.85               & 0                   & Absent             & CE                              & 50.71          & 24.63       & 0.78                                     \\
A2                                      & Self-supervised (standard training)                          & 70.85               & 0                   & Absent             & $\cos(\mathcal{T},\mathcal{S})$ & 54.48          & 23.20       & 0.78                                     \\
A3                                      & \multicolumn{1}{l}{Supervised (TRADES adversarial training)} & 59.88               & 25.89               & Absent             & CE                              & 54.86          & 27.17       & 0.94                                     \\
A4                                      & Self-supervised (standard training)                          & 70.85               & 0                   & Present            & CE                              & 57.51          & 24.10       & 0.18   \\
                         \bottomrule
\end{tabular}}
\vspace{-0.2cm}
\end{table*}

\emph{Student model performance improves by matching the following during distillation:
\begin{enumerate}
\vspace{-0.3cm}
    \item Training objectives of the teacher and the ideal goals of the student, 
    \vspace{-0.2cm}
    \item Pretraining and linear probe training objectives of the student.
\end{enumerate}}

Here, the ideal goal of the student depends on the downstream task, which is clean (or standard) accuracy in standard training, and clean and robust accuracy in adversarial training. On the other hand, the training objective of the standard self-supervised trained teacher is to achieve invariance to augmentations of the same image when compared to augmentations of other images.

We now explain the intuition behind the above hypotheses and empirically justify the same by considering several distillation settings involving standard and adversarial, supervised and self-supervised trained teacher models in \cref{tab:motiv-SD,tab:motiv-AD}. The results are presented on CIFAR-100 \citep{krizhevsky2009learning} with WideResNet-34-10 \citep{zagoruyko2016wide} architecture for both teacher and student. The standard self-supervised model is trained using SimCLR \citep{simclr}. Contrary to a typical knowledge distillation setting where a cross-entropy loss is also used \citep{hinton2015distilling}, all the experiments presented involve the use of only self-supervised losses for distillation (cosine similarity between representations), and labels are used only during linear probing. Adversarial self-supervised distillation in \cref{tab:motiv-AD} is performed using a combination of distillation loss on natural samples and smoothness loss on adversarial samples as shown in \cref{eq:f} \citep{deacl}. A randomly initialized trainable projector is used at the output of student backbone in S5 of \cref{tab:motiv-SD} and A4 of \cref{tab:motiv-AD}. Here, the training loss is considered in the projection space of the student ($\mathcal{S}_p$) rather than the feature space ($\mathcal{S}_f$).

\textbf{1. Matching the training objectives of teacher with the ideal goals of the student:} 
Consider task-A to be the teacher's training task, and task-B to be the student's downstream task or its ideal goal. The representations in deeper layers (last few layers) of the teacher are more tuned to its training objective, and the early layers contain a lot more information than what is needed for this task \citep{bordes2022guillotine}. Thus, features specific to task-A are dominant or replicated in the final feature layer, and other features that may be relevant to task-B are sparse. When a similarity based distillation loss is enforced on such features, higher importance is given to matching the task-A dominated features, and the sparse features which may be important for task-B are suppressed further in the student \citep{addepalli2023feature}. On the other hand, when the student's task matches with the teacher's task, a similarity based distillation loss is very effective in transferring the necessary representations to the student, since they are predominant in the final feature layer. Thus, matching the training objective of the teacher with the ideal goals of the student should improve downstream performance.

To test this hypothesis, we first consider the standard training of a student model, using either a self-supervised or supervised teacher, in \cref{tab:motiv-SD}. One can note that in the absence of a projector, the drop in student accuracy w.r.t. the respective teacher accuracy is $6\%$ with a self-supervised teacher (S1), and $<0.5\%$ with a supervised teacher (S3). To ensure that our observations are not a result of the $10\%$ difference in teacher accuracy between S1 and S3, we present results and similar observations with a supervised sub-optimally trained teacher in S4. Thus, a supervised teacher is significantly better than a self-supervised teacher for distilling representations specific to a given task. This justifies the hypothesis that, \emph{student performance improves by matching the training objectives of the teacher and the ideal goals of the student}.

We next consider adversarial training of a student, using either a standard self-supervised teacher, or a supervised adversarially trained (TRADES) teacher, in \cref{tab:motiv-AD}. Since the TRADES model is more aligned with the ideal goals of the student, despite its lower clean accuracy, the clean and robust accuracy of the student are better than those obtained using a standard self-supervised model as a teacher (A3 vs. A1). This further justifies the first hypothesis. 

\textbf{2. Matching the pretraining and linear probe training objectives of the student:}
For a given network, aligning the pretraining task with downstream task results in better performance since the matching of tasks ensures that the required features are predominant, and they are easily used by, for e.g., an SVM or a linear classifier trained over it \citep{addepalli2023feature}. In context of distillation, since the features of the student are trained by enforcing similarity based loss w.r.t. the teacher, we hypothesize that enforcing similarity w.r.t. the teacher is the best way to learn the student classifier as well. To illustrate this, let's consider task-A to be the teacher pretraining task, and task-B to be the downstream task or ideal goal of the student. As discussed above, the teacher's features are aligned to task-A and these are transferred effectively to the student. The features related to task-B are suppressed in the teacher and are further suppressed in the student. As the features specific to a given task become more sparse, it is harder for an SVM (or a linear) classifier to rely on that feature, although it important for classification \citep{addepalli2023feature}. Thus, training a linear classifier for task-B is more effective on the teacher when compared to the student. The linear classifier of the teacher in effect amplifies the sparse features, allowing the student to learn them more effectively. Thus, training a classifier on the teacher and distilling it to the student is better than training a classifier directly on the student.

We now provide empirical evidence in support of this hypothesis. To align pretraining with linear probing, we perform linear probing on the teacher model, and further train the student by maximizing the cosine similarity between the logits of the teacher and student. This boosts the student accuracy by 3.6\%, in \cref{tab:motiv-SD} (S2 vs. S1) and by $3.8\%$ in \cref{tab:motiv-AD} (A2 vs. A1). The projector isolates the representations of the student from the training loss, as indicated by the lower similarity between the student and teacher at feature space when compared to that at the projector (in S5 and A4), and prevents overfitting of the student to the teacher training objective. This makes the student robust to the misalignment between the teacher training objective and ideal goals of the student, and also to the mismatch in student pretraining and linear probing objectives, thereby improving student performance, as seen in \cref{tab:motiv-SD} (S5 vs. S1) and \cref{tab:motiv-AD} (A4 vs. A1).  

\begin{figure*}
\centering
\includegraphics[width=0.7\linewidth]{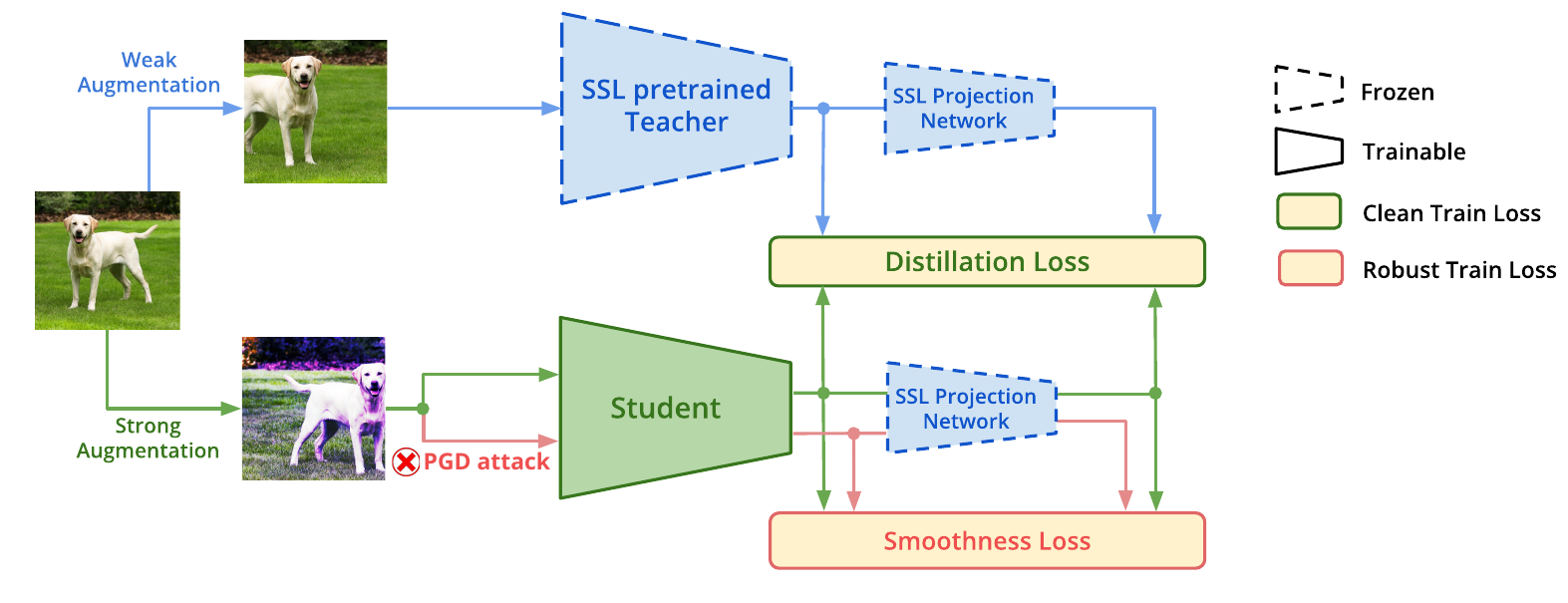}
 
\caption{\textbf{Proposed approach (ProFeAT):} The student is trained using a distillation loss on clean samples using supervision from an SSL pretrained teacher, and a smoothness loss to enforce adversarial robustness (details of exact loss formulation is presented in \cref{sec:proposed_method}). A frozen pretrained projection layer is used at the teacher and student to prevent overfitting to the clean distillation loss. The use of strong augmentations at the student increases attack diversity, while weak augmentations at the teacher reduce the training complexity.}
\label{fig:profeat_diag}
\vspace{-0.5cm}
\end{figure*}

\subsection{ProFeAT: Projected Feature Adversarial Training}
\vspace{-0.1cm}
We present details on the proposed approach ProFeAT, illustrated in \cref{fig:profeat_diag}. Firstly, a teacher model is trained using a self-supervised training algorithm such as SimCLR \citep{simclr}, which is also used as an initialization for the student for better convergence. 

\textbf{Use of Projection Layer:} As discussed in \cref{subsec:motiv-proj}, to overcome the impact of the inherent misalignment between the training objective of the teacher and the ideal goals of the student, and the mismatch between the pretraining and linear probing objectives, we propose to use a projection head at the output of the student backbone. As noted in \cref{tab:motiv-SD} (S5 vs. S1) and \cref{tab:motiv-AD} (A4 vs. A1), even a randomly initialized projection head improves performance. Most self-supervised pretraining methods use similarity based losses at the output of a projection head for training \citep{simclr,moco,byol,simsiam,barlow}, resulting in a projected space where similarity has been enforced during pretraining, thus giving higher importance to the key dimensions. We therefore propose to reuse this pretrained projection head for both teacher and student and freeze it during training to prevent convergence to an identity mapping.

\textbf{Defense Loss:} As is common in adversarial training literature \citep{zhang2019theoretically,deacl}, we use a combination of loss on clean samples and smoothness loss to enforce adversarial robustness in the student model. Since the loss on clean samples utilizes supervision from the self-supervised pretrained teacher, it is enforced at the outputs of the respective projectors of the teacher and student as discussed above. The goal of the second loss is merely to enforce local smoothness in the loss surface of the student, and is enforced in an unsupervised manner \citep{zhang2019theoretically,deacl}. Thus, it is ideal to enforce this loss at the feature space of the student network, since these representations are directly used for downstream applications. 
While the ideal locations for the clean and adversarial losses are the projected and feature spaces respectively, we find that such a loss formulation is hard to optimize, resulting in either a non-robust model, or collapsed representations as shown in \cref{tab:abl-beta-AD5}. We thus use a complimentary loss as a regularizer in the respective spaces. This results in a combination of losses at the feature and projector spaces as shown below: 

\setlength\abovedisplayskip{0pt}
\setlength\belowdisplayskip{0pt}

\vspace{-0.1cm}

\begin{align}
\label{eq:fp}
    &\mathcal{L}_{fp} = - \sum_i \cos\big(\mathcal{T}_{fp}(x_i), \mathcal{S}_{fp}(x_i)\big) + \beta \cos\big(\mathcal{S}_{fp}(x_i), \mathcal{S}_{fp}(\tilde{x}_i)\big) \\
\label{eq:f}
    &\mathcal{L}_f = - \sum_i \cos\big(\mathcal{T}_{f}(x_i), \mathcal{S}_{f}(x_i)\big) +
     \beta \cos\big(\mathcal{S}_{f}(x_i), \mathcal{S}_{f}(\tilde{x}_i)\big) \\
\label{eq:defense-loss}
    &\mathcal{L}_\mathrm{ProFeAT} = \frac{1}{2} \big(\mathcal{L}_{fp} + \mathcal{L}_f\big) \\
\label{eq:attack-loss}
    &\tilde{x}_i = \argmin_{\tilde{x}_i: ||\tilde{x}_i-x_i||_\infty \leq \varepsilon} \cos\big(\mathcal{T}_{fp}(x_i), \mathcal{S}_{fp}(\tilde{x}_i)\big) + 
    \cos\big(\mathcal{S}_{f}(x_i), \mathcal{S}_{f}(\tilde{x}_i)\big)
\end{align}

Here, $L_{pf}$ and $L_f$ are the defense losses enforced at the projector and feature spaces, respectively. $\mathcal{T}_{pf_i} = \left(\mathcal{T}_{p} \circ \mathcal{T}_{f}\right)(x_i)$ is the composition of the projection layer $\mathcal{T}_{p}$ on the feature backbone $\mathcal{T}_{f}$ of the teacher $\mathcal{T}$ for a clean input $x_i$. $\tilde{\mathcal{S}} = \mathcal{S}(\tilde{x})$ where $\tilde{x}$ is the adversarial input and $\mathcal{S}$ represents student representation (similar subscript notations follow for the student). The first term in \cref{eq:fp,eq:f} represents the \textbf{Distillation loss} (\cref{fig:profeat_diag}), whereas the second term corresponds to the \textbf{Smoothness loss} at the respective layers of the student, and is weighted by a hyperparameter $\beta$ that controls the robustness-accuracy trade-off in the downstream model. The overall loss $\mathcal{L}_\mathrm{ProFeAT}$ (\cref{eq:defense-loss}) is minimized during training.

\textbf{Attack generation:} The attack used during training is generated by a maximizing a combination of losses at both projector and feature spaces as shown in \cref{eq:attack-loss}. Since the projector space is primarily used for enforcing similarity with the teacher, we minimize the cosine similarity between the teacher and student representations for attack generation. Since the feature space is primarily used for enforcing local smoothness in the loss surface of the student, we utilize the unsupervised formulation that minimizes similarity between representations of clean and adversarial samples at the student. 

\textbf{Augmentations:} Standard supervised and self-supervised training approaches are known to benefit from the use of strong data augmentations such as AutoAugment \citep{cubuk2018autoaugment}. However, such augmentations, which distort the low-level features of images, are known to deteriorate the performance of adversarial training \citep{rice2020overfitting,gowal2020uncovering}. \citet{addepalli2022efficient} attribute the poor performance to the larger domain shift between the augmented train and unaugmented test set images, in addition to the increased complexity of the adversarial training task, which overpower the superior generalization attained due to the use of diverse augmentations. Although these factors influence adversarial training in the self-supervised regime as well, we hypothesize that the need for better generalization is higher in self-supervised training, since the pretraining task is not aligned with the ideal goals of the student, making it important to use strong augmentations. However, it is also important to ensure that the training task is not too complex. We thus propose to use a combination of weak and strong augmentations as inputs to the teacher and student respectively, as shown in \cref{fig:profeat_diag}. From \cref{fig:attack-restarts} we note that, the use of strong augmentations results in the generation of more diverse attacks, resulting in a larger drop when differently augmented images are used across different restarts of a PGD 5-step attack. The use of weak augmentations at the teacher imparts better supervision to the student, reducing the training complexity. 
\vspace{-0.1cm}
\section{Experiments and Results}
\vspace{-0.1cm}
We first present an empirical evaluation of the proposed method, followed by several ablation experiments to understand the role of each component individually. 
Details on the datasets, training and compute are presented in \cref{sec:datasets,sec:training_compute}. 

We now present the experimental results comparing the proposed approach ProFeAT with respect to several existing self-supervised adversarial training approaches \citep{chen2020adversarial,rocl,acl,advcl,deacl,zhang2019theoretically} by freezing the feature extractor and performing linear probing (LP) using cross-entropy loss on clean samples. To ensure a fair comparison, the same is done for the supervised AT method TRADES \citep{zhang2019theoretically} as well. 
The results are presented on CIFAR-10 and CIFAR-100 datasets \cite{krizhevsky2009learning}, and on ResNet-18 \cite{he2016deep} and WideResNet-34-10 (WRN-34-10) \cite{zagoruyko2016wide} architectures, as common in the adversarial research community. The results of existing methods on ResNet-18 are as reported by \citet{deacl}. Since DeACL \citep{deacl} also uses a teacher-student architecture, we reproduce their results using the same teacher as our method, and report the same as ``DeACL (\textcolor{black}{Our Teacher})''. Since existing methods do not report results on larger architectures like WideResNet-34-10, we compare our results only with the best performing method (DeACL) and a recent approach DynACL \citep{dynacl}. These results are not reported in the respective papers, hence we run them using the official code. 

The \textcolor{black}{Robust Accuracy (RA)} in the SOTA comparison table is presented against AutoAttack (RA-AA) \citep{croce2020reliable} which is widely used as a reliable benchmark for robustness evaluation \citep{croce2021robustbench}. In other tables, we also present robust accuracy against the GAMA attack (RA-G) \citep{sriramanan2020gama} which is known to be a competent and a reliable estimate of AutoAttack while being significantly faster to evaluate. We additionally present results against a 20-step PGD attack (RA-PGD20) \citep{madry-iclr-2018} in the SOTA comparison table (\cref{tab:main-LP}), although it is a significantly weaker attack. A larger difference between RA-PGD20 and RA-AA indicates that the loss surface is more convoluted due to which weaker attacks are unsuccessful, yielding a false sense of robustness \citep{athalye2018obfuscated}. Thus this difference serves as a check for verifying the extent of gradient masking \citep{papernot2017practical, tramer2017ensemble}. Therefore, in order to compare true robustness between any two defenses, accuracy against RA-AA or RA-G should be considered, while RA-PGD20 should not be considered. 
The accuracy on clean or natural samples is denoted as SA, which stands for Standard (Clean) Accuracy.

\vspace{-0.1cm}

\subsection{Comparison with the state-of-the-art}

\begin{table*}
\centering
\caption{\textbf{SOTA comparison:} Standard Linear Probing performance (\%) on CIFAR-10 and CIFAR-100 datasets on ResNet-18 and WideResNet-34-10 models. Mean and standard deviation across 3 reruns are reported for the best baseline DeACL \citep{deacl} and the proposed approach ProFeAT. Standard Accuracy (SA), Robust Accuracy against PGD-20 attack (RA-PGD20) and AutoAttack (RA-AA) reported.}
\resizebox{1.0\linewidth}{!}{
\label{tab:main-LP}
\begin{tabular}{lllllll}
\toprule
    \multirow{2}{*}{\textbf{Method}}                & \multicolumn{3}{c}{\textbf{CIFAR-10}}      & \multicolumn{3}{c}{\textbf{CIFAR-100}}     \\
                    \cmidrule{2-7}
                    & \textbf{SA} & \textbf{\textcolor{black}{RA-PGD20}}      & \textbf{\textcolor{black}{RA-AA}}  & \textbf{SA} & \textbf{\textcolor{black}{RA-PGD20}}      & \textbf{\textcolor{black}{RA-AA}}  \\
                    \midrule
                    \multicolumn{7}{c}{\textbf{ResNet-18}}                                                  \\
                    \midrule
                    \midrule
Supervised (TRADES)  & 83.74          & 49.35       & 47.60       & 59.07          & 26.22       & 23.14       \\
\midrule
AP-DPE              & 78.30          & 18.22       & 16.07       & 47.91          & 6.23        & 4.17        \\
RoCL                 & 79.90          & 39.54       & 23.38       & 49.53          & 18.79       & 8.66        \\
ACL                 & 77.88          & 42.87       & 39.13       & 47.51          & 20.97       & 16.33       \\
AdvCL               & 80.85          & 50.45       & 42.57       & 48.34          & 27.67       & 19.78       \\
DynACL              & 77.41          & -           & 45.04       & 45.73          & -           & 19.25       \\
DynACL++              & 79.81          & -           & 46.46       & 52.26          & -           & 20.05       \\
DeACL (Reported)    & 80.17          & 53.95       & 45.31       & 52.79          & 30.74       & 20.34       \\
DeACL (\textcolor{black}{Our Teacher})  & 80.05\std{0.29}    & 52.97\std{0.08} & \textbf{48.15}\std{0.05} & 51.53\std{0.30}     & 30.92\std{0.21} & 21.91\std{0.13} \\
ProFeAT  (\textbf{Ours})           & \textbf{81.68}\std{0.23}    & 49.55\std{0.16} & 47.02\std{0.01} & \textbf{53.47}\std{0.10}    & 27.95\std{0.13} & \textbf{22.61}\std{0.14}  \\
\midrule
                    \multicolumn{7}{c}{\textbf{WideResNet-34-10}}                                           \\
                    \midrule
                    \midrule
Supervised (TRADES)  & 85.50          & 54.29       & 51.59       & 59.87          & 28.86       & 25.72       \\
\midrule
 \textcolor{black}{DynACL++}       &       \textcolor{black}{80.97}	& \textcolor{black}{48.28} &	\textcolor{black}{45.50}	& \textcolor{black}{52.60}	& \textcolor{black}{23.42} &	\textcolor{black}{20.58}     \\
DeACL        &       83.83\std{0.20}	&57.09\std{0.06}&	48.85\std{0.11}	&52.92\std{0.35}	&32.66\std{0.08}&	23.82\std{0.07}     \\
ProFeAT   (\textbf{Ours})            & \textbf{87.62}\std{0.13}    & 54.50\std{0.17} & \textbf{51.95}\std{0.19} & \textbf{61.08}\std{0.18}    & 31.96\std{0.08} & \textbf{26.81}\std{0.11} \\
\bottomrule
\end{tabular}}
\end{table*}

\begin{table}[t]
\color{black}
\centering
\caption{\textbf{Performance with additional evaluations:} Performance (\%) of DeACL (best baseline) and ProFeAT (Ours) on CIFAR-10 and CIFAR-100 datasets with WideResNet-34-10 architecture. The model is first pretrained using the respective self-supervised adversarial training algorithm, and further we compute the standard accuracy (SA) and robust accuracy against GAMA (RA-G) using several methods such as standard linear probing (LP), training a 2 layer MLP head (MLP), and performing KNN in the feature space with $k=10$. The proposed method achieves improvements over the baseline across all evaluation methods.}
\setlength\tabcolsep{12pt}
\resizebox{0.85\linewidth}{!}{
\label{tab:supple-knn}
\begin{tabular}{lcccccc}
\toprule
\multirow{2}{*}{\textbf{Method}} & \multicolumn{2}{c}{\textbf{LP Eval}} & \multicolumn{2}{c}{\textbf{MLP Eval}} & \multicolumn{2}{c}{\textbf{KNN Eval} \boldmath$(k=10)$} \\
\cmidrule{2-7}
 & \textbf{SA} & \textbf{RA-G} & \textbf{SA} & \textbf{RA-G} & \textbf{SA} & \textbf{RA-G} \\
\midrule
\multicolumn{7}{c}{\textbf{CIFAR-10}}                                                                                     \\
\midrule
DeACL                & 83.60            & 49.62              & 85.66             & 48.74               & 87.00             & 54.58               \\
ProFeAT (\textbf{Ours})             & \textbf{87.44}            & \textbf{52.24}              & \textbf{89.37}             & \textbf{50.00}               & \textbf{87.38}             & \textbf{55.77}               \\
\midrule
\multicolumn{7}{c}{\textbf{CIFAR-100}}                                                                                    \\
\midrule
DeACL                & 52.90            & 24.66              & 55.05             & 22.04               & 56.82             & 31.26               \\
ProFeAT (\textbf{Ours})              & \textbf{61.05}            & \textbf{27.41}              & \textbf{63.81}             & \textbf{26.10}               & \textbf{58.09}             & \textbf{32.26}         \\
\bottomrule
\end{tabular}
}
\vspace{-0.3cm}
\end{table}

\cref{tab:main-LP} presents the standard linear probing results of the proposed method ProFeAT in comparison to several SSL-AT baseline approaches. 
The proposed approach obtains superior robustness-accuracy trade-off when compared to the best performing baseline method DeACL, with $\sim3{-}3.5\%$ gains in both robust and clean accuracy on CIFAR-10 dataset and similar gains in robustness on CIFAR-100 dataset with WideResNet-34-10 architecture. We obtain significant gains of $\sim8\%$ on the clean accuracy on CIFAR-100. With ResNet-18 architecture, ProFeAT achieves competent robustness-accuracy trade-off when compared to DeACL on CIFAR-10 dataset, and obtains $\sim2\%$ higher clean accuracy alongside improved robustness on CIFAR-100 dataset. We notice significantly less gradient masking, indicated by a lower value of (RA-PGD20 $-$ RA-AA), for the proposed approach compared to all other baselines across all settings, indicating a reliable attack generation even in the absence of ground truth labels. Overall, the proposed approach significantly outperforms all the existing baselines, especially for larger model capacities (WRN-34-10), with improved results on smaller models (ResNet-18). Additionally, we obtain superior results when compared to the supervised AT method TRADES as well, at higher model capacities. 

\textbf{Performance comparison with other evaluation methods:} We also present results of additional evaluation methods to evaluate the performance of the pretrained backbone in \cref{tab:supple-knn}. We note that the proposed method achieves improvements over the baseline across all evaluation methods. Since the training of classifier head in LP and MLP is done using standard training and not adversarial training, the robust accuracy reduces as the number of layers increases (from linear to 2-layers), and the standard accuracy improves. The standard accuracy of KNN is better than the standard accuracy of LP for the baseline, indicating that the representations are not linearly separable. Whereas, as is standard, for the proposed approach, LP standard accuracy is higher than that obtained using KNN. The adversarial attack used for evaluating the robust accuracy using KNN is generated using GAMA attack on a linear classifier. The attack is suboptimal since it is not generated by using the evaluation process (KNN), and thus the robust accuracy against such an attack is higher.

\textbf{Transfer learning:} To evaluate the transferrability of the learned robust representations, we compare the proposed approach with the best baseline DeACL in \cref{tab:main-TL} under standard linear probing (LP). We consider transfer from CIFAR-10/100 to STL-10 \citep{coates2011analysis}. When compared to DeACL, the clean accuracy is $\sim4-10\%$ higher on CIFAR-10 and $\sim 1.7-6\%$ higher on CIFAR-100. We also obtain $3-5\%$ higher robust accuracy when compared to DeACL on CIFAR-100, and higher improvements over TRADES.  We also present transfer learning results using lightweight adversarial full finetuning (AFF) to STL-10 and Caltech-101 \citep{li_andreeto_ranzato_perona_2022} in \cref{tab:supple-stl10}. We defer the details on the process of adversarial full-finetuning and the selection criteria for the transfer datasets to \cref{sec:training_compute}.
The transfer is performed on WRN-34-10 model that is pretrained on CIFAR-10/100. As shown in \cref{tab:supple-stl10}, the proposed method outperforms DeACL by a significant margin. Note that by using merely 25 epochs of adversarial full-finetuning, the proposed method achieves improvements of around 4\% on CIFAR-10 and 11\% on CIFAR-100 when compared to the linear probing accuracy presented in \cref{tab:main-TL}, highlighting the practical utility of the proposed method. The AFF performance of the proposed approach is better than that of a supervised TRADES pretrained model as well.

\begin{table}
\begin{minipage}{0.49\linewidth}
\centering
\caption{\textbf{Transfer Learning with Linear Probing:} Standard LP Performance (\%) for transfer learning from CIFAR-10/100 to STL-10 dataset on ResNet-18 and WRN-34-10 model. ProFeAT (Ours) outperforms both DeACL and the supervised TRADES model.}
\vspace{-0.1cm}
\setlength\tabcolsep{10pt}
\resizebox{1.0\linewidth}{!}{
\label{tab:main-TL}
\begin{tabular}{lcccc}
\toprule
\multirow{3}{*}{\textbf{Method}} & \multicolumn{2}{c}{\begin{tabular}[x]{@{}c@{}}\small{\textbf{CIFAR-10}}\\\small{$\rightarrow$ \textbf{STL-10}}\end{tabular}} & \multicolumn{2}{c}{\begin{tabular}[x]{@{}c@{}}\small{\textbf{CIFAR-100}}\\\small{$\rightarrow$ \textbf{STL-10}}\end{tabular}} \\
\cmidrule{2-5}
& \textbf{SA} & \textbf{RA-AA} & \textbf{SA} & \textbf{RA-AA} \\
\midrule
\multicolumn{5}{c}{\textbf{ResNet-18}} \\

                  \midrule
Supervised        & 54.70 & 22.26 & 51.11 & 19.54                \\
DeACL             & 60.10 & 30.71 & 50.91 & 16.25                \\
ProFeAT    & \textbf{64.30} & \textbf{30.95}       & \textbf{52.63} & \textbf{20.55}       \\
\midrule 
\multicolumn{5}{c}{\textbf{WideResNet-34-10}} \\
\midrule 
Supervised        & 67.15 & 30.49 & \textbf{57.68} & 11.26                \\
DeACL             & 66.45 & 28.43 & 50.59 & 13.49                \\
ProFeAT    & \textbf{69.88} & \textbf{31.65} & 56.68 & \textbf{19.46}  \\

\bottomrule
\end{tabular}}
\end{minipage}
\hfill
\begin{minipage}{0.49\linewidth}
\centering
\caption{\textbf{Transfer Learning with Adv. Full-Finetuning:} AFF performance (\%) using TRADES algorithm for 25 epochs, when transferred from CIFAR-10/100 to STL-10 and Caltech-101 dataset on WideResNet-34-10 model.}
\vspace{-0.15cm}
\setlength\tabcolsep{10pt}
\resizebox{1.0\linewidth}{!}{
\label{tab:supple-stl10}
\begin{tabular}{lcccc}
\toprule
\multirow{3}{*}{\textbf{Method}} & \textbf{SA} & \textbf{RA-G} & \textbf{SA} & \textbf{RA-G}             \\
\cmidrule{2-5}
 & \multicolumn{2}{c}{\begin{tabular}[x]{@{}c@{}}\small{\textbf{CIFAR-10}}\\\small{$\rightarrow$ \textbf{STL-10}}\end{tabular}} & \multicolumn{2}{c}{\begin{tabular}[x]{@{}c@{}}\small{\textbf{CIFAR-100}}\\\small{$\rightarrow$ \textbf{STL-10}}\end{tabular}} \\
\midrule
Supervised & 64.58 & 32.78            & 64.22 & 31.01            \\
DeACL & 61.65 & 28.34 & 60.89 & 30.00            \\
ProFeAT & \textbf{74.12} & \textbf{36.04} & \textbf{68.77} & \textbf{31.23}   \\
\midrule
\multicolumn{1}{l}{} & \multicolumn{2}{c}{\begin{tabular}[x]{@{}c@{}}\small{\textbf{CIFAR-10}}\\\small{$\rightarrow$ \textbf{Caltech-101}}\end{tabular}} & \multicolumn{2}{c}{\begin{tabular}[x]{@{}c@{}}\small{\textbf{CIFAR-100}}\\\small{$\rightarrow$ \textbf{Caltech-101}}\end{tabular}} \\
\midrule
Supervised  & 62.46 & 39.40 & \textbf{64.97} & 41.02              \\
DeACL & 62.65 & 39.18 & 61.01 & 39.09           \\
ProFeAT              & \textbf{66.11} & \textbf{42.12}    & 64.16 & \textbf{41.25}       \\
\bottomrule
\end{tabular}}
\end{minipage}
\end{table}

\begin{table}[t]
\centering
\caption{\textbf{Efficiency of Self-supervised adversarial training (CIFAR-100, WRN-34-10):} Performance (\%) using lesser number of attack steps (2 steps) when compared to the standard case (5 steps) during adversarial training. Clean/ Standard Accuracy (SA) and robust accuracy against GAMA attack (\textcolor{black}{RA-G}) and AutoAttack(\textcolor{black}{RA-AA}) are reported. The proposed approach is stable at lower attack steps as well, while being better than both TRADES \citep{zhang2019theoretically} and DeACL \citep{deacl}.}
\vspace{-0.1cm}
\setlength\tabcolsep{20pt}
\resizebox{1.0\linewidth}{!}{
\label{tab:supp-abl-attack-steps}
\begin{tabular}{lcccccc}
\toprule
\multirow{2}{*}{\textbf{Method}} & \multicolumn{3}{c}{\textbf{2-step PGD attack}} & \multicolumn{3}{c}{\textbf{5-step PGD attack}} \\
\cmidrule{2-7}
                                          & \textbf{SA}    & \textbf{RA-G}    & \textbf{RA-AA}    & \textbf{SA} & \textbf{RA-G} & \textbf{RA-AA} \\
                                          \midrule
Supervised (TRADES)                                & \textbf{60.80}          & 24.49          & 23.99          & \textbf{61.05}       & 25.87       &      25.77   \\
DeACL                                & 51.00          & 24.89          & 23.45          & 52.90       & 24.66       & 23.92 \\
ProFeAT (\textbf{Ours}) & 60.43 & 26.90 & \textbf{26.23} & \textbf{61.05} & 27.41 & \textbf{26.89} \\
\bottomrule
\end{tabular}
}
\vspace{-0.3cm}
\end{table}

\textbf{Efficiency of self-supervised adversarial training}
Similar to prior works \citep{deacl}, the proposed approach uses 5-step PGD based optimization for attack generation during adversarial training. In \cref{tab:supp-abl-attack-steps}, we present results with lesser optimization steps (2 steps). The proposed approach is stable and obtains similar results even by using 2-step attack. Even in this case, the clean and robust accuracy of the proposed approach is significantly better than the baseline approach DeACL \citep{deacl}, and also outperforms the supervised TRADES model \citep{zhang2019theoretically}. We compare the computational aspects of the existing baselines and the proposed method in more detail in \cref{app:computational complexity}.

\subsection{Ablations}

We now present some of the ablation experiments to gain further insights into the proposed method, and defer more in-depth ablation results to \cref{sec:additional_results} due to space constraints.

\textbf{Effect of each component of the proposed approach:} We study the impact of each component of the ProFeAT in Table-\ref{tab:supple-components}, and make the following observations based on the results:
\vspace{-0.2cm}
\begin{itemize}

    \item \textit{Projector}: A key component of the proposed method is the introduction of the projector. We observe significant gains in clean accuracy ($\sim 5\%$) by introducing the projector along with defense losses at the feature and projection spaces (E1 vs. E2). The importance of the projector is also evident by the fact that removing the projector from the proposed defense results in a large drop ($5.7\%$) in clean accuracy (E9 vs. E5). 
    We observe a substantial improvement of $9.2\%$ in clean accuracy when the projector is introduced in the presence of the proposed augmentation strategy (E3 vs. E7), which is significantly higher than the gains obtained by introducing the same in the baseline DeACL ($4.76\%$, E1 vs. E2). Further ablations on the projector are provided in \cref{subsec:supp-abl-proj}.
    
    \item \textit{Augmentations}: The proposed augmentation strategy improves robustness across all settings. Introducing the proposed strategy in the baseline improves its robust accuracy by $2.47\%$ (E1 vs. E3). Moreover, the importance of the proposed strategy is also evident from the fact that in the absence of the same, there is a $4.48\%$ drop in SA and $\sim 2\%$ drop in RA-G (E9 vs. E6). Further, when combined with other components as well, the proposed augmentation strategy shows good improvements (E4 vs. E5, E2 vs. E7, E6 vs. E9). Detailed ablation on the respective augmentations used for the teacher and student model can be found in \cref{subsec:supp-abl-augs}.
    \item \textit{Attack loss}: The proposed attack objective is designed to be consistent with the proposed defense strategy, where the goal is to enforce smoothness at the student in the feature space and similarity with the teacher in the projector space. 
    The impact of the attack loss in feature space can be seen in combination with the proposed augmentations, where we observe an improvement of $2.5\%$ in clean accuracy alongside notable improvements in robust accuracy (E3 vs. E5). However, in presence of projector, the attack results in only marginal robustness gains, possibly because the clean accuracy is already high (E9 vs. E7). More detailed ablations on the attack loss in provided in \cref{subsec:suppl-attack-loss}.
    \item \textit{Defense loss}: We do not introduce a separate column for defense loss as it is applicable only in the presence of the projector. We show the impact of the proposed defense losses in the last two rows (E8 vs. E9). The proposed defense loss improves the clean accuracy by $1.4\%$ and robust accuracy marginally. \cref{subsec:suppl-defense-loss} provides further insights on various defense loss formulations and their impact on the proposed method.
\end{itemize}

\begin{table}[t]
\color{black}
\centering
\caption{\textbf{Ablations on ProFeAT (CIFAR-100, WRN-34-10): } Performance (\%) by enabling different components of the proposed approach. A tick mark in the Projector column means that a frozen pretrained projector is used for the teacher and student, with the defense loss being enforced at the feature and projector as shown in \cref{eq:defense-loss}. E1 represents the baseline or DeACL defense, and E9 represents the proposed defense or ProFeAT. E8*:Defense loss applied only at the projector. SA: Standard Accuracy, RA-G: Robust Accuracy against GAMA attack.}
\setlength\tabcolsep{15pt}
\resizebox{0.95\linewidth}{!}{
\label{tab:supple-components}
\begin{tabular}{cccccc}
\toprule
\textbf{Ablation} & \textbf{Projector}   & \textbf{Augmentations} & \textbf{Attack loss} & \textbf{SA} & \textbf{RA-G}  \\
\midrule
E1  & \multicolumn{1}{l}{} &                        & \multicolumn{1}{l}{} & 52.90 & 24.66          \\
E2  & \checkmark                    &                        & \multicolumn{1}{l}{} & 57.66 & 25.04          \\
E3  & \multicolumn{1}{l}{} & \multicolumn{1}{c}{\checkmark}  & \multicolumn{1}{l}{} & 52.83 & 27.13          \\
E4  & \multicolumn{1}{l}{} &                        & \checkmark                    & 51.80 & 24.77          \\
E5  & \multicolumn{1}{l}{} & \multicolumn{1}{c}{\checkmark}  & \checkmark                    & 55.35 & \textbf{27.86} \\
E6  & \checkmark                    &                        & \checkmark                    & 56.57 & 25.29          \\
E7  & \checkmark                    & \multicolumn{1}{c}{\checkmark}  & \multicolumn{1}{l}{} & \textbf{62.01} & 26.89          \\
E8  & \checkmark*                & \multicolumn{1}{c}{\checkmark}  & \checkmark                    & 59.65 & 26.90          \\
E9  & \checkmark                    & \multicolumn{1}{c}{\checkmark}  & \checkmark                    & 61.05 & 27.41         \\
\bottomrule
\end{tabular}}
\end{table}

\begin{figure}
\centering
\begin{minipage}{0.45\linewidth}
\centering
\includegraphics[width=\textwidth]{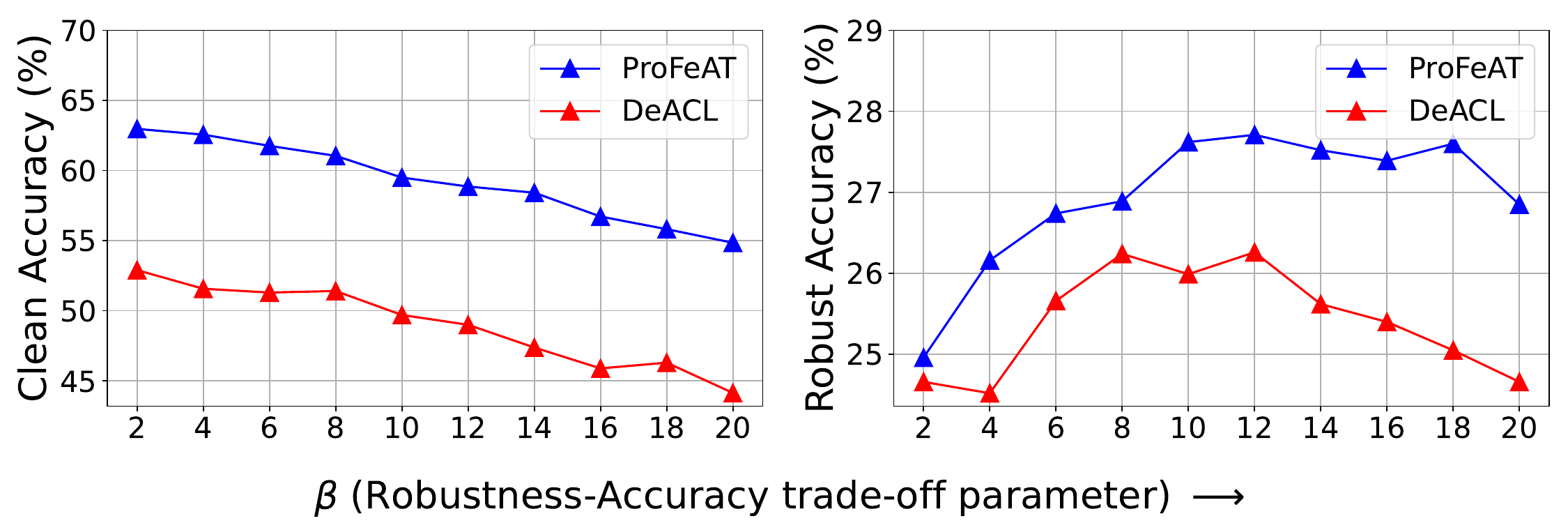}

\caption{\small{Performance of ProFeAT when compared to DeACL \citep{deacl} across variation in the robustness-accuracy trade-off parameter $\beta$ on CIFAR-100 dataset with WRN-34-10 architecture.}}
\label{fig:beta-var}
\end{minipage}
\hfill
\begin{minipage}{0.53\linewidth}
\centering
\includegraphics[width=\textwidth]{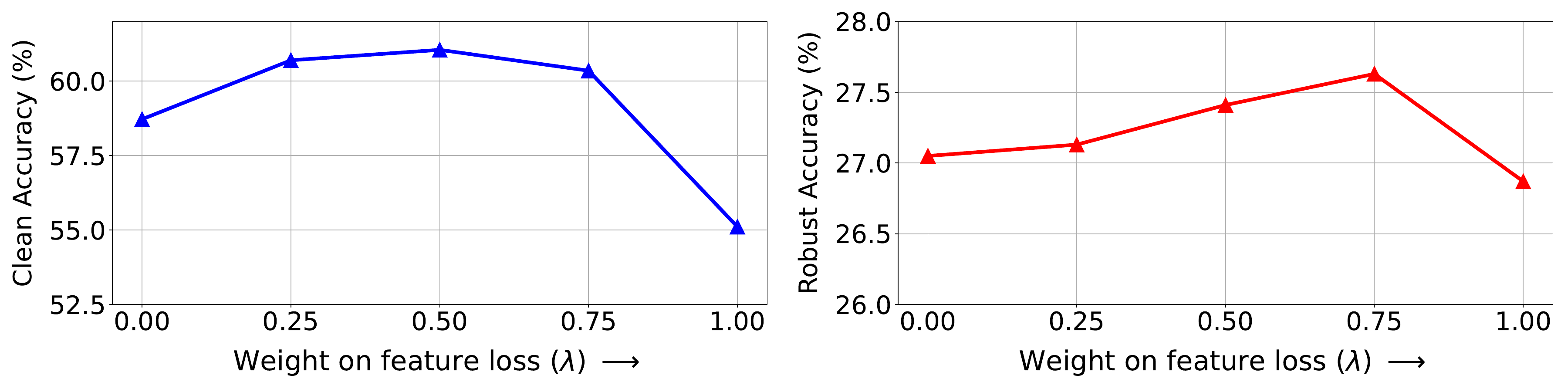}
\caption{\small{Performance ($\%$) of ProFeAT by varying the weight $\lambda$ between the defense losses at the feature and projector. The performance is stable across the range $\lambda \in [0.25,0.75]$. We thus fix the value of $\lambda$ to 0.5 in the proposed approach.}}
\label{fig:loss_w}
\end{minipage}
\end{figure}

\begin{table}
\centering
\caption{\textbf{Performance across different architecture types and sizes:} Standard Linear Probing performance (\%) of DeACL (Baseline) and ProFeAT (Ours) across different architectures on CIFAR-100. ViT-B/16 uses Imagenet-1K trained SSL teacher for training, while the SSL teacher in all other cases is trained on the CIFAR-100. \textcolor{black}{SA: Standard Accuracy, RA-AA: Robust Accuracy against AutoAttack.}}
\setlength{\tabcolsep}{10pt}
\resizebox{0.9\linewidth}{!}{
\label{tab:abl-arch}
\begin{tabular}{lccccc}
\toprule
\multirow{2}{*}{\textbf{Method}} & \multirow{2}{*}{\textbf{\#params (M)}} & \multicolumn{2}{c}{\textbf{DeACL}}               & \multicolumn{2}{c}{\textbf{ProFeAT (Ours)}}      \\  
\cmidrule{3-6}
                  &                                                                                  & \textbf{\textcolor{black}{SA}} & \textbf{\textcolor{black}{RA-AA}} & \textbf{\textcolor{black}{SA}} & \textbf{\textcolor{black}{RA-AA}} \\ \midrule
ResNet-18         & 11.27                                                                            & 51.53             & 21.91      & \textbf{53.47}             & \textbf{22.61}      \\ 
ResNet-50         & 23.50                                                                            & 53.30             & 23.00      & \textbf{59.34}             &\textbf{25.86}      \\
WideResNet-34-10  & 46.28                                                                            & 52.92             & 23.82      & \textbf{61.08}             & \textbf{26.81}      \\
ViT-B/16          & 85.79                                                                            & 61.34             & 17.49       & \textbf{65.08}             & \textbf{21.52}      \\ \bottomrule
\end{tabular}}
\vspace{0.1cm}
\end{table}

\textbf{Performance across different model architectures:}
We report performance of the proposed method ProFeAT and the best baseline DeACL on diverse architectures including Vision transformers \citep{dosovitskiy2020image} on the CIFAR-100 dataset in \cref{tab:abl-arch}. ProFeAT consistently outperforms DeACL in both clean and robust accuracy across various model architectures. An explanation behind the mechanism for successful scaling to larger datasets can be found in \cref{sec:suppl-scaling}.

\textbf{Robustness-Accuracy trade-off:}
We present results across variation in the robustness-accuracy trade-off parameter $\beta$ (\cref{eq:fp,eq:f}) in \cref{fig:beta-var}. Both robustness and accuracy of the proposed method are significantly better than DeACL across all values of $\beta$. 

\textbf{Weighting of defense losses at the feature and projector:}
In the proposed approach, the defense losses are equally weighted between the feature and projector layers as shown in \cref{eq:defense-loss}. In \cref{fig:loss_w}, we present results by varying the weighting $\lambda$ between the defense losses at the feature ($\mathcal{L}_{f}$) and projector ($\mathcal{L}_{fp}$) layers: $\mathcal{L}_\mathrm{ProFeAT} = \lambda \cdot \mathcal{L}_f + (1-\lambda) \cdot \mathcal{L}_{fp}$, where $L_{fp}$ and $L_f$ are given by \cref{eq:fp,eq:f}, respectively. It can be noted that the two extreme cases of $\lambda=0$ and $\lambda=1$ result in a drop in clean accuracy, with a larger drop in the case where the loss is enforced only at the feature layer. The robust accuracy shows lesser variation across different values of $\lambda$. Thus, the overall performance is stable over the range $\lambda \in [0.25,0.75]$, making the default setting of $\lambda=0.5$ a suitable option. 

\section{Conclusion}
\vspace{-0.2cm}

To summarize, in this work, we bridge the performance gap between supervised and self-supervised adversarial training approaches, specifically for large capacity models. We utilize a teacher-student setting \citep{deacl} where a standard self-supervised trained teacher is used to provide supervision to the student. Due to the inherent misalignment between the teacher training objective and the ideal goals of the student, we propose to use a projection layer to prevent the network from overfitting to the standard SSL trained teacher. We present a detailed analysis on the use of projection layer in distillation to justify our method. We additionally propose appropriate attack and defense losses in the feature and projector spaces alongside the use of weak and strong augmentations for the teacher and student respectively, to improve the attack diversity while maintaining low training complexity. The proposed approach obtains significant gains over existing self-supervised adversarial training methods, specifically for large models, demonstrating its scalability. 

\bibliography{main}
\bibliographystyle{tmlr}

\newpage
\appendix

\section*{Appendix}

\section{Background: Supervised Adversarial Defenses}

Following the demonstration of adversarial attacks by \citet{intriguing-iclr-2014}, there have been several attempts of defending Deep Networks against them. Early defenses proposed intuitive methods that introduced non-differentiable or randomized components in the network to thwart gradient-based attacks \citep{buckman2018thermometer,ma2018characterizing,s.2018stochastic,xie2018mitigating,song2018pixeldefend}. While these methods were efficient and easy to implement, \citet{athalye2018obfuscated} proposed adaptive attacks which successfully broke several such defenses by replacing the non-differentiable components with smooth differentiable approximations, and by taking an expectation over the randomized components. Adversarial Training \citep{madry-iclr-2018,zhang2019theoretically} was the most successful defense strategy that withstood strong white-box \citep{croce2020reliable,sriramanan2020gama}, black-box \citep{andriushchenko2019square,chen2020rays} and adaptive attacks \citep{athalye2018obfuscated,tramer2020adaptive} proposed over the years. PGD (Projected Gradient Descent) based adversarial training \citep{madry-iclr-2018} involves maximizing the cross-entropy loss to generate adversarial attacks, and further minimizing the loss on the adversarial attacks for training. Another successful supervised Adversarial Training based defense was TRADES \citep{zhang2019theoretically}, where the Kullback-Leibler (KL) divergence between clean and adversarial samples was minimized along with the cross-entropy loss on clean samples for training. Adjusting the weight between the losses gave a flexible trade-off between the clean and robust accuracy of the trained model. Although these methods have been robust against several attacks, it has been shown that the sample complexity of adversarial training is large \citep{schmidt2018adversarially}, and this increases the training and annotation costs needed for adversarial training.

\color{black}

\section{Mechanism behind Scaling to Larger Datasets} 
\label{sec:suppl-scaling}
For a sufficiently complex task, a scalable approach results in better performance on larger models given enough data. Although the task complexity of adversarial self-supervised learning is high, the gains in prior approaches are marginal with an increase in model size, while the proposed method results in significantly improved performance on larger capacity models (\cref{tab:main-LP}). We discuss the key factors that result in better scalability below:
\begin{itemize}
    \item As discussed in \cref{subsec:motiv-proj}, a mismatch between training objectives of the teacher and ideal goals of the student causes a drop in student performance. This primarily happens because of the overfitting to the teacher training task. As model size increases, the extent of overfitting increases. The use of a projection layer during distillation alleviates the impact of this overfitting and allows the student to retain more generic features that are useful for the downstream robust classification objective. Thus, a projection layer is more important for larger model capacities where the extent of overfitting is higher.
    \item Secondly, as the model size increases, there is a need for higher amount of training data for achieving better generalization. The proposed method has better data diversity as it enables the use of more complex data augmentations in adversarial training by leveraging supervision from weak augmentations at the teacher.
\end{itemize}

\section{Details on Datasets}
\label{sec:datasets}
We compare the performance of the proposed approach ProFeAT with existing methods on the benchmark datasets  CIFAR-10 and CIFAR-100 \citep{krizhevsky2009learning}, that are commonly used for evaluating the adversarial robustness of models \citep{croce2021robustbench}. Both datasets consist of RGB images of dimension $32 \times 32$. CIFAR-10 consists of 50,000 images in the training set and 10,000 images in the test set, with the images being divided equally into 10 classes - airplane, automobile, bird, cat, deer, dog, frog, horse, ship and truck. CIFAR-100 dataset is of the same size as CIFAR-10, with images being divided equally into 100 classes. Due to the larger number of classes, there are only 500 images per class in CIFAR-100, making it a more challenging dataset when compared to CIFAR-10. 

\begin{table}[]
\caption{\textbf{Total number of Forward (FP) or Backward (BP) Propagations during training} of the proposed approach when compared to prior works. Distillation based approaches - ProFeAT and DeACL require significantly lesser compute when compared to prior methods, and are only more expensive than supervised adversarial training.}
\setlength\tabcolsep{12pt}
\resizebox{1.0\linewidth}{!}{
\label{tab:compute}
\begin{tabular}{lccccc}
\toprule
\multirow{2}{*}{\textbf{Method}}    & \multirow{2}{*}{\textbf{\#Epochs}} & \multirow{2}{*}{\textbf{\#Attack steps}} & \multicolumn{3}{c}{\textbf{\#FP or BP}} \\
\cmidrule{4-6}
 & & & \textbf{AT} & \textbf{Teacher model} & \textbf{Total} \\
\midrule
Supervised (TRADES) & 110       & 10              & 1210              & -                             & 1210             \\ 
AP-DPE                       & 450       & 10              & 4950              & -                             & 4950             \\ 
RoCL                         & 1000      & 5               & 6000              & -                             & 6000             \\ 
ACL                          & 1000      & 5               & 12000              & -                             & 12000             \\ 
AdvCL                        & 1000      & 5               & 12000              & 1000                          & 13000             \\ 
DynACL                        & 1000      & 5               & 12000              & -                          & 12000             \\ 
DynACL++                        & 1025      & 5               & 12300              & -                          & 12300             \\ 
DeACL                        & 100       & 5               & 700              & 1000                          & 1700             \\
ProFeAT (\textbf{Ours})                      & 100       & 5               & 800              & 1000                          & 1800             \\ \bottomrule
\end{tabular}}
\end{table}

\section{Details on Training and Compute}
\label{sec:training_compute}
\textbf{Model Architecture:} We report the key comparisons with existing methods on two of the commonly considered model architectures in the literature of adversarial robustness \citep{pang2020bag,zhang2019theoretically,rice2020overfitting,croce2021robustbench} - ResNet-18 \citep{he2016deep} and WideResNet-34-10 \citep{zagoruyko2016wide}. Although most existing methods for self-supervised adversarial training report results only on ResNet-18 \citep{deacl,advcl}, we additionally consider the WideResNet-34-10 architecture to demonstrate the scalability of the proposed approach to larger model architectures. We perform the ablation experiments on the CIFAR-100 dataset with WideResNet-34-10 architecture, which is a very challenging setting in self-supervised adversarial training, to be able to better distinguish between different variations adopted during training. 

\textbf{Training Details:} The self-supervised training of the teacher model is performed for 1000 epochs with the SimCLR algorithm \citep{simclr} similar to prior work \citep{deacl}. We utilize the \texttt{solo-learn} \footnote{\url{https://github.com/vturrisi/solo-learn}} GitHub repository for this purpose. For the SimCLR SSL training, we tune and use a learning rate of 1.5 with SGD optimizer, a cosine schedule with warmup, weight decay of $1e{-}5$ and train the backbone for 1000 epochs with other hyperparameters kept as default as in the repository.

The self-supervised adversarial training of the feature extractor using the proposed approach is performed for 100 epochs using SGD optimizer with a weight decay of $3e{-}4$, cosine learning rate with $10$ epochs of warm-up, and a maximum learning rate of $0.5$. We consider the standard $\ell_\infty$ based threat model \cite{deacl,advcl} for attack generation during both training and evaluation, i.e., $||\tilde{x}_i - x_i||_\infty \leq \varepsilon$, where $\tilde{x}$ is the adversarial version of the clean sample $x_i$. The value of $\varepsilon$ is set to $8/255$ for both CIFAR-10 and CIFAR-100 datasets, as is standard in literature \citep{madry-iclr-2018,zhang2019theoretically}. A 5-step PGD attack is used for attack generation during training with a step size of $2/255$. We fix the value of $\beta$, the robustness-accuracy trade-off parameter (Ref: \cref{eq:fp,eq:f} in the main paper) to $8$ in all our experiments, unless specified otherwise. 

\textbf{Details on Linear Probing:} To evaluate the performance of the learned representations, we perform standard linear probing by freezing the adversarially pretrained backbone as discussed in \cref{sec:prelims} of the main paper. We use a class-balanced validation split consisting of $1000$ images from the train set and perform early-stopping during training based on the performance on the validation set. The training is performed for $25$ epochs with a step learning rate schedule where the maximum learning rate is decayed by a factor of $10$ at epoch $15$ and $20$. The learning rate is chosen amongst the following settings --- $\{0.1, 0.05, 0.1, 0.5, 1, 5\}$, with SGD optimizer, and the weight decay is fixed to $2e{-}4$. 
The same evaluation protocol is used for the best baseline DeACL \citep{deacl} as well as the proposed approach ProFeAT, for both in-domain and transfer learning settings. 

\textbf{Details on Transfer Learning:} To perform transfer learning using Adversarial Full-Finetuning, a robustly pretrained base model is used as an initialization, which is then adversarially finetuned using TRADES \cite{zhang2019theoretically} for $25$ epochs. After finetuning, the model is evaluated using linear probing on the same transfer dataset. We consider STL-10 \cite{coates2011analysis} and Caltech-101 \cite{li_andreeto_ranzato_perona_2022} as the transfer datasets. Caltech-101 contains 101 object classes and 1 background class, with 2416 samples in the train set and 6728 samples in the test set. The number of samples per class range from 17 to 30, and thus is a suitable dataset to highlight the practical importance of lightweight finetuning of an adversarial self-supervised pretrained representations for low-data regime.

\textbf{Compute:} The following Nvidia GPUs have been used for performing the experiments reported in this work - V100, A100, and A6000. Each of the experiments are run either on a single GPU, or across 2 GPUs based on the complexity of the run and GPU availability. Uncertainty estimates in the main SOTA table (\cref{tab:main-LP} of the main paper) were obtained on the same machine and with same GPU configuration to ensure reproducibility. For $100$ epochs of single-precision (FP32) training with a batch size of $256$, the proposed approach takes $\sim$8 hours and  $\sim$16GB of GPU memory on a single A100 GPU for WideResNet-34-10 model on CIFAR-100.

\begin{table}[]
\centering
\caption{\textbf{Floating Point Operations per Second (GFLOPS) and latency per epoch during training} of the proposed approach ProFeAT when compared to the baseline DeACL for ResNet-18 and WideResNet-34-10 models. The computational overhead during training is marginal with the addition of the projection layer, and reduces further for larger capacity models.}
\resizebox{1.0\linewidth}{!}{
\label{tab:flops}
\begin{tabular}{lcccccc}
\toprule
\multirow{2}{*}{} & \multicolumn{3}{c}{\textbf{ResNet-18}}           & \multicolumn{3}{c}{\textbf{WideResNet-34-10}}    \\
\cmidrule{2-7}
                  & \textbf{GFLOPS} & \textbf{Time/epoch } & \textbf{\#Params(M)} & \textbf{GFLOPS} & \textbf{Time/epoch} &\textbf{ \#Params(M)} \\ \midrule
DeACL             & 671827     & 51s         & 11.27        & 6339652    & 4m 50s      & 46.28        \\ 
ProFeAT (\textbf{Ours})    & 672200     & 51s         & 11.76        & 6340197    & 4m 50s      & 46.86        \\
\midrule
\% increase       & 0.056      & 0.00        & 4.35         & 0.009      & 0.00        & 1.25         \\ \bottomrule
\end{tabular}
}
\vspace{-0.2cm}
\end{table}

\section{Computational Complexity}
\label{app:computational complexity}
In terms of compute, both the proposed method ProFeAT and DeACL \citep{deacl} lower the overall computational cost when compared to prior approaches. This is because self-supervised training in general requires larger number of epochs (1000) to converge \citep{simclr,moco} when compared to supervised learning (\textless100). Prior approaches like RoCL \citep{rocl}, ACL \citep{acl} and AdvCL \citep{advcl} combine the contrastive training objective of SSL approaches and the adversarial training objective. Thus, these methods require larger number of training epochs (1000) for the adversarial training task, which is already computationally expensive due to the requirement of generating multi-step attacks during training. ProFeAT and DeACL use a SSL teacher for training and thus, the adversarial training is more similar to supervised training, requiring only 100 epochs. In \cref{tab:compute}, we present the approximate number of forward and backward propagations for each algorithm, considering both pretraining of the auxiliary network used and the training of the main network. It can be noted that the distillation based approaches - ProFeAT and DeACL require significantly lesser compute when compared to prior methods, and are only more expensive than supervised adversarial training. In \cref{tab:flops}, we present the FLOPS required during training for the proposed approach and DeACL. One can observe that there is a negligible increase in FLOPS compared to the baseline approach.

\begin{table*}
\centering
\caption{\textbf{Ablation on Projector training configuration (CIFAR-100, WRN-34-10): } Performance (\%) using variations in projector (proj.) initialization (init.) and trainability. \textcolor{black}{SA: Standard Accuracy, RA-G: Robust accuracy against GAMA attack.}}
\setlength\tabcolsep{6pt}
\resizebox{1.0\linewidth}{!}{
\label{tab:abl-proj}
\begin{tabular}{lllllccc}
\toprule
\textbf{Ablation} & \textbf{Student proj.} & \textbf{Proj. init. (Student)} & \textbf{Teacher proj} & \textbf{Proj. init. (Teacher)} & \textbf{SA} & \textbf{\textcolor{black}{RA-G}} \\
\midrule
AP1         & Absent                     & -                                 & Absent                     & -                                 & 55.35 & 27.86       \\
AP2         & Trainable                  & Random                            & Absent                     & -                                 & 63.07 & 26.57       \\
AP3         & Frozen                     & Pretrained                   & Absent                     & -                                 & 40.43 & 22.23       \\
AP4         & Trainable                  & Pretrained                   & Absent                     & -                                 & 62.89 & 26.57       \\
AP5         & Trainable                  & Random (common)                   & Trainable                  & Random (common)                   & 53.43 & 27.23       \\
AP6         & Trainable                  & Pretrained (common)          & Trainable                  & Pretrained (common)          & 54.60 & 27.41       \\
AP7         & Trainable                  & Pretrained                   & Frozen                     & Pretrained                   & 58.18 & 27.73  \\
\textbf{Ours}         & Frozen                     & Pretrained                   & Frozen                     & Pretrained                   & 61.05 & 27.41       \\

\bottomrule
\end{tabular}}
\end{table*}

\section{Additional Results}
\label{sec:additional_results}
\begin{table}[t]
\centering
\caption{\textbf{Ablation on Projector architecture (CIFAR-100, WRN-34-10): } Performance (\%) obtained by varying the projector configuration (config.) and architecture (arch.). A non-linear projector effectively reduces the gap in clean accuracy between the teacher and student. A bottleneck architecture for the projector is worse than other variants. \textcolor{black}{SA: Standard Accuracy, RA-G: Robust Accuracy against GAMA attack.}}
\resizebox{1.0\linewidth}{!}{
\label{tab:supp-abl-proj-config}
\begin{tabular}{cllccccc}
\toprule
\textbf{Ablation}   & \textbf{Projector config.} & \textbf{Projector arch.} & \textbf{Teacher \textcolor{black}{SA}} & \textbf{Student \textcolor{black}{SA}} & \textbf{Drop in \textcolor{black}{SA}} & \textbf{\% Drop in \textcolor{black}{SA}} & \textbf{\textcolor{black}{RA-G}} \\
\midrule
APA1          & No projector               & -                        & 70.85                 & 55.35                 & 15.50                 & 21.88 & \textbf{27.86}       \\
APA2          & Linear layer               & 640-256                  & 68.08                 & 53.35                 & 14.73                 & 21.64 & 27.47       \\
\textbf{Ours} & 2 Layer MLP                & 640-640-256              & 70.85                 & 61.05                 & 9.80                  & 13.83 & 27.41       \\
APA3          & 3 Layer MLP                & 640-640-640-256          & 70.71                 & 60.37                 & 10.34                 & 14.62 & 27.37       \\
APA4          & 2 Layer MLP                & 640-640-640              & 69.88                 & 61.24                 & 8.64                  & 12.36 & 27.36       \\
APA5          & 2 Layer MLP                & 640-2048-640             & \textbf{70.96}                 & \textbf{61.76}                 & 9.20                  & 12.97 & 26.66       \\
APA6          & 2 Layer MLP                & 640-256-640              & 69.37                 & 57.87                 & 11.50                 & 16.58 & 27.56   \\
\bottomrule
\end{tabular}}
\vspace{-0.2cm}
\end{table}

\begin{table}
\begin{minipage}{0.56\linewidth}
\centering
\vspace{0.2cm}
\setlength\tabcolsep{10pt}
\resizebox{1.0\linewidth}{!}{
\begin{tabular}{ccccc}
\toprule
\textbf{Ablation} & \textbf{Teacher} & \textbf{Student} & \textbf{SA} & \textbf{\textcolor{black}{RA-G}}  \\
\midrule
AG1         & PC               & PC               & 56.57 & 25.29       \\
AG2         & \textcolor{black}{AuAu}               & \textcolor{black}{AuAu}               & 60.76 & 27.21       \\
AG3         & PC1              & PC2              & 56.95 & 25.39       \\
AG4           & \textcolor{black}{AuAu}1              & \textcolor{black}{AuAu}2              & 59.51 & \textbf{28.15}       \\
AG5           & \textcolor{black}{AuAu}               & PC               & 57.28 & 26.14       \\
Ours & PC               & \textcolor{black}{AuAu}               & \textbf{61.05} & 27.41      \\
\bottomrule
\end{tabular}}
\caption{\small{\textbf{Ablation on Augmentations used (CIFAR-100, WRN-34-10):} Performance (\%) using different augmentations for the teacher and student. (PC: Pad+Crop, \textcolor{black}{AuAu}: AutoAugment). \textcolor{black}{Standard Accuracy (SA) and Robust accuracy (RA-G) reported.}}}
\label{tab:abl-augs}
\end{minipage}
\hfill
\begin{minipage}{0.4\linewidth}
\centering
\centering
\includegraphics[width=0.98\linewidth]{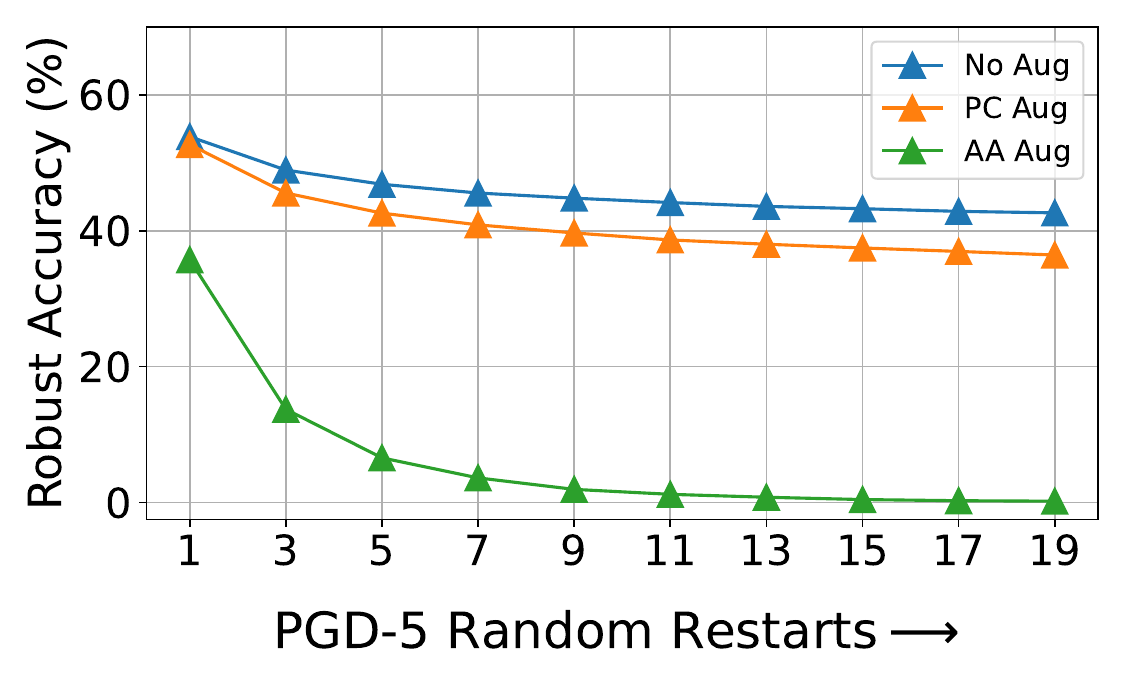}
\captionof{figure}{\small{Robust accuracy of a supervised TRADES model across random restarts of PGD 5-step attack (CIFAR-100, WRN-34-10).}}
\label{fig:attack-restarts}
\end{minipage}
\end{table}
\subsection{More ablations on the Projector}
\label{subsec:supp-abl-proj}

\textbf{Training configuration of the Projector:} We present ablations using different configurations of the projection layer in \cref{tab:abl-proj}. As also discussed in \cref{subsec:motiv-proj} of the main paper, we observe a large boost in clean accuracy when a random (or pretrained) trainable projection layer is introduced to the student (AP2/ AP4 vs. AP1 in \cref{tab:abl-proj}). While the use of pretrained frozen projection head only for the student degrades performance considerably (AP3), the use of the same for both teacher and student (Ours) yields a optimal robustness-accuracy trade-off across all variations. The use of a common trainable projection head for both teacher and student results in collapsed representations at the projector output (AP5, AP6), yielding results similar to the case where projector is not used for both teacher and student (AP1). This issue is overcome when the pretrained projector is trainable only for the student (AP7). 

\begin{table}
\centering
\caption{\textbf{Ablation on Attack Loss (CIFAR-100, WRN-34-10): } Performance (\%) with variations in attack loss at feature (feat.) and projector (proj.). While the proposed defense is stable across several variations in the attack loss, minimizing a combination of both losses $\cos(\mathcal{T},\mathcal{S})$ and $\cos(\mathcal{S},\mathcal{S})$ gives the best robustness-accuracy trade-off. \textcolor{black}{SA: Standard Accuracy, RA-G: Robust Accuracy against GAMA.}}
\setlength\tabcolsep{15pt}
\resizebox{0.9\linewidth}{!}{
\label{tab:supple-abl-attack}
\begin{tabular}{ccccc}
\toprule
\textbf{Ablation}   & \textbf{Attack @ feat.} & \textbf{Attack @ proj.} & \textbf{\textcolor{black}{SA}} & \textbf{\textcolor{black}{RA-G}} \\
\midrule
AT1           & $\cos(\mathcal{T},\mathcal{S})$               & $\cos(\mathcal{T},\mathcal{S})$               & 60.84 & 26.78                         \\
AT2           & $\cos(\mathcal{S},\mathcal{S})$               & $\cos(\mathcal{S},\mathcal{S})$               & 61.30 & 26.75                         \\
AT3           & $\cos(\mathcal{T},\mathcal{S})$               & $\cos(\mathcal{S},\mathcal{S})$               & 60.69 & \textbf{27.44}                         \\
AT4           & $\cos(\mathcal{S},\mathcal{S})$               & -                      & 61.62 & 26.62                         \\
AT5           & -                      & $\cos(\mathcal{S},\mathcal{S})$               & 61.09 & 27.00                         \\
AT6           & $\cos(\mathcal{T},\mathcal{S})$               & -                      & \textbf{62.01} & 26.89               \\
AT7           & -                      & $\cos(\mathcal{T},\mathcal{S})$               & 61.18 & 27.24    \\
Ours & $\cos(\mathcal{S},\mathcal{S})$               & $\cos(\mathcal{T},\mathcal{S})$               & 61.05 & 27.41                         \\
\bottomrule
\end{tabular}
}
\end{table}
\begin{table}
\centering
\caption{\textbf{Ablation on Defense Loss (CIFAR-100, WRN-34-10): } Performance (\%) with variations in training loss at feature (feat.) and projector (proj.). ``clean" denotes the cosine similarity between representations of teacher and student on clean samples. ``adv" denotes the cosine similarity between representations of the corresponding clean and adversarial samples either at the output of student $(\mathcal{S},\mathcal{S})$ or between the teacher and student $(\mathcal{T},\mathcal{S})$. \textcolor{black}{SA: Standard Accuracy, RA-G: Robust accuracy against GAMA.}}
\setlength\tabcolsep{10pt}
\resizebox{0.8\linewidth}{!}{
\label{tab:abl:defense}
\begin{tabular}{ccccc}
\toprule
\textbf{Ablation} & \textbf{Loss @ feat.} & \textbf{Loss @ proj.} & \textbf{SA} & \textbf{RA-G}  \\
\midrule
AD1         & clean + adv$(\mathcal{S},\mathcal{S})$       & -                       & 55.35 & \textbf{27.86}         \\
AD2         & -                       & clean + adv$(\mathcal{S},\mathcal{S})$       & 59.65 & 26.90        \\
AD3         & clean + adv$(\mathcal{S},\mathcal{S})$       & clean                   & 61.69 & 26.40         \\
AD4         & clean + adv$(\mathcal{S},\mathcal{S})$       & adv$(\mathcal{S},\mathcal{S})$               &    49.59 &25.35          \\
AD5         & adv$(\mathcal{S},\mathcal{S})$               & clean                   & 59.72 & 1.38                 \\
AD6         & adv$(\mathcal{S},\mathcal{S})$               & clean + adv$(\mathcal{S},\mathcal{S})$       & 59.22 & 26.50                \\
AD7         & clean                   & clean + adv$(\mathcal{S},\mathcal{S})$       & 62.24 & 25.97                \\
AD8         & clean + adv$(\mathcal{S},\mathcal{S})$       & clean + adv$(\mathcal{T},\mathcal{S})$       & 63.85 & 23.91                \\
AD9        & clean + adv$(\mathcal{T},\mathcal{S})$       & clean + adv$(\mathcal{T},\mathcal{S})$       & \textbf{65.34} & 22.40 \\
Ours         & clean + adv$(\mathcal{S},\mathcal{S})$       & clean + adv$(\mathcal{S},\mathcal{S})$       & 61.05 & 27.41        \\
\bottomrule
\end{tabular}}
\end{table}

\begin{table}
\centering
\caption{\textbf{Failure of AD5 (\cref{tab:abl:defense}) defense loss for SSL-AT training of WRN-34-10 model on CIFAR-100:} Using clean and adversarial loss exclusively at projector and feature space respectively, results in an unstable optimization problem, giving either a non-robust model or collapsed representations at the end of training. As shown below, a lower value of $\beta$ (the robustness-accuracy trade-off parameter) results in a non-robust model, while higher $\beta$ results in the learning of collapsed representations.}
\label{tab:abl-beta-AD5}
\begin{tabular}{ccc}

\toprule
$\boldsymbol{\beta}$ & \textbf{Standard Accuracy (SA)}  & \textbf{\textcolor{black}{Robust Accuracy against GAMA (RA-G)}} \\ 
\midrule
1                    & \textbf{67.34}   & 0.46 \\ 
5                    & 51.99   & 0.71 \\ 
10                   & 31.34  & 7.81 \\ 
50                   & 11.59   & 2.55 \\ 
100                  & 8.23    & \textbf{2.61} \\ 
\bottomrule
\end{tabular}
\vspace{-0.1cm}
\end{table}

\textbf{Architecture of the Projector:} In the proposed approach, we use the following 2-layer MLP projector architecture for both SSL pretraining of the teacher and adversarial training of the student: (1) \textbf{ResNet-18: 512-512-256}, and (2) \textbf{WideResNet-34-10: 640-640-256}. In \cref{tab:supp-abl-proj-config}, we present results using different configurations and architectures of the projector. Firstly, the use of a linear projector (APA2) is similar to the case where projector is not used for student training (APA1), with $\sim21\%$ drop in clean accuracy of the student with respect to the teacher. This improves to $12-17\%$ when a non-linear projector is introduced (APA3-APA6 and Ours). The use of a 2-layer MLP (Ours) is marginally better than the use of a 3-layer MLP (APA3) in terms of clean accuracy of the student. The accuracy of the student is stable across different architectures of the projector (Ours, APA4, APA5). However, the use of a bottleneck architecture (APA6) results in a higher drop in clean accuracy of the student.

\subsection{Augmentation for the Student and Teacher}
\label{subsec:supp-abl-augs}
We present ablation experiments to understand the impact of different augmentations used for the teacher and student separately in \cref{tab:abl-augs}. The base method (AG1) uses common Pad and Crop (PC) augmentation for both teacher and student. By using more complex augmentations ---AutoAugment followed by Pad and Crop (denoted as \textcolor{black}{AuAu} in the table), there is a significant improvement in both clean and robust accuracy. By using separate augmentations for the teacher and student, there is an improvement in the case of PC (AG3), but a drop in clean accuracy accompanied by better robustness in case of \textcolor{black}{AuAu}. Finally by using a mix of both \textcolor{black}{AuAu} and PC at the student and teacher respectively (Ours), we obtain improvements in both clean and robust accuracy, since the former improves attack diversity (shown in \cref{fig:attack-restarts}), while the latter makes the training task easier.

\subsection{Attack Loss}
\label{subsec:suppl-attack-loss}

For performing adversarial training using the proposed approach, attacks are generated by minimizing a combination of cosine similarity based losses as shown in \cref{eq:attack-loss} of the main paper. This includes an unsupervised loss at the feature representations of the student and another loss between the representations of the teacher and student at the projector. As shown in \cref{tab:supple-abl-attack}, we obtain a better robustness-accuracy trade-off by using a combination of both losses rather than by using only one of the two losses, due to better diversity and strength of attack. These results also demonstrate that the proposed method is not very sensitive to different choices of attack losses. 

\subsection{Defense Loss}
\label{subsec:suppl-defense-loss}

We present ablation experiments across variations in training loss at the feature space and the projection head in \cref{tab:abl:defense}. In the proposed approach (Ours), we introduce a combination of clean and robust losses at both feature and projector layers, as shown in \cref{eq:defense-loss} of the main paper. By introducing the loss only at the features (AD1), there is a considerable drop in clean accuracy as seen earlier, which can be recovered by introducing the clean loss at the projection layer (AD3). Instead, when only the robust loss is introduced at the projection layer (AD4), there is a large drop in clean accuracy confirming that the need for projection layer is mainly enforcing the clean loss. When the combined loss is enforced only at the projection head (AD2), the accuracy is close to that of the proposed approach, with marginally lower clean and robust accuracy. Enforcing only adversarial loss in the feature space, and only clean loss in the projector space is a hard optimization problem, and this results in a non-robust model (AD5). As shown in \cref{tab:abl-beta-AD5}, even by increasing $\beta$ in AD5, we do not obtain a robust model, rather, there is a representation collapse. Thus, as discussed in \cref{subsec:motiv-proj} of the main paper, it is important to introduce the adversarial loss as a regularizer in the projector space as well (AD6). Enforcing only one of the two losses at the feature space (AD6 and AD7) also results in either inferior clean accuracy or robustness. Finally from AD8 and AD9 we note that the robustness loss is better when implemented as a smoothness constraint on the representations of the student, rather than by matching representations between the teacher and student. Overall, the proposed approach (Ours) results in the best robustness-accuracy trade-off. 

\subsection{Accuracy of the self-supervised teacher model}
\begin{table}[t]
\centering
\caption{\textbf{Ablations on the accuracy of the teacher SSL model (CIFAR-100, WRN-34-10): } Performance (\%) obtained by varying the number of epochs for which the standard self-supervised teacher model is pretrained. Improvements in accuracy of the teacher result in corresponding gains in both standard and robust accuracy of the student. \textcolor{black}{SA: Standard Accuracy, RA-G: Robust Accuracy against GAMA.}}
\setlength\tabcolsep{10pt}
\resizebox{1.0\linewidth}{!}{
\label{tab:supp-abl-teacher-training}
\begin{tabular}{cccccc}
\toprule
\textbf{\#Epochs of PT} & \textbf{Teacher \textcolor{black}{SA}} & \textbf{Student \textcolor{black}{SA}} & \textbf{Drop in \textcolor{black}{SA}} & \textbf{\% Drop in \textcolor{black}{SA}} & \textbf{\textcolor{black}{RA-G}} \\
\midrule
100                      & 55.73                 & 49.37                 & 6.36                  & 11.41 & 20.86       \\
200                      & 65.43                 & 56.16                 & 9.27                  & 14.17 & 24.15       \\
500                      & 69.27                 & 59.62                 & 9.65                  & 13.93 & 26.75       \\
1000                     & \textbf{70.85}                 & \textbf{61.05}                 & 9.80                  & 13.83 & \textbf{27.41} \\
\bottomrule
\end{tabular}
}
\end{table}

The self-supervised teacher model is obtained using 1000 epochs of SimCLR \citep{simclr} training in all our experiments. We now study the impact of training the teacher model for lesser number of epochs. As shown in \cref{tab:supp-abl-teacher-training}, as the number of teacher training epochs reduces, there is a drop in the accuracy of the teacher, resulting in a corresponding drop in the clean and robust accuracy of the student model. Thus, the performance of the teacher is crucial for training a better student model. 

\label{subsec:suppl-teacher-train-ep}

\subsection{Self-supervised training algorithm of the teacher}
\label{subsec:suppl-ssl-algo}
In the proposed approach, the teacher is trained using the popular self-supervised training algorithm SimCLR \citep{simclr}, similar to prior works \citep{deacl}. In this section, we study the impact of using different algorithms for the self-supervised training of the teacher and present results in \cref{tab:supp-abl-ssl-pt}. In order to ensure consistency across different SSL methods, we use a \textit{random trainable} projector (2-layer MLP with both hidden and output dimensions of 640) for training the student and do not employ any projection head for the pretrained frozen teacher. While the default teacher trained using SimCLR was finetuned across hyperparameters, we utilize the default hyperparameters from the \texttt{solo-learn} Github repository for this table, and thus present SimCLR also without tuning for a fair comparison. For uniformity, we report all results with $\beta = 8$ (the robustness-accuracy trade-off parameter). From \cref{tab:supp-abl-ssl-pt}, we note that in most cases, the clean accuracy of the student increases as the accuracy of the teacher improves, while the robust accuracy does not change much. We note that this table merely shows that the proposed approach can be effectively integrated with several base self-supervised learning algorithms for the teacher model. However, it does not present a fair comparison across different SSL pretraining algorithms, since the ranking on the final performance of the student would change if the pretraining SSL algorithms were used with appropriate hyperparameter tuning. 

\begin{table}[t]
\centering
\caption{\textbf{Ablations on the algorithm used for training the self-supervised teacher model (CIFAR-100, WRN-34-10): } Performance (\%) of the proposed approach by varying the pretraining algorithm of the teacher model. A random trainable projector is used for training the student model, to maintain uniformity in projector architecture across all methods. \textcolor{black}{SA: Standard Accuracy, RA-G: Robust Accuracy against GAMA attack.}}
\setlength\tabcolsep{15pt}
\resizebox{0.9\linewidth}{!}{
\label{tab:supp-abl-ssl-pt}
\begin{tabular}{lccc}
\toprule
\multicolumn{1}{c}{\textbf{Method (Teacher training)}} & \textbf{Teacher \textcolor{black}{SA}} & \textbf{Student \textcolor{black}{SA}} & \textbf{\textcolor{black}{RA-G}} \\
\midrule
SimCLR                                        & 67.98                 & 62.20 & 26.13                     \\
SimCLR (tuned)                                & 70.85                 & 63.07 & 26.57                     \\
BYOL                                          & \textbf{72.97}                 & 63.19 & \textbf{26.82}                     \\
Barlow Twins                                  & 67.74                 & 60.69 & 24.48                     \\
SimSiam                                       & 68.60                 & 63.46 & 26.69                     \\
MoCoV3                                        & 72.48                 & \textbf{65.57} & 26.65                     \\
DINO                                          & 68.75                 & 60.61 & 24.80         \\
\bottomrule
\end{tabular}
}
\end{table}

\end{document}